\documentclass[10pt,twocolumn,letterpaper]{article}

\usepackage{iccv}
\usepackage{times}
\usepackage{epsfig}
\usepackage{graphicx}
\usepackage{amsmath, bm}
\usepackage{amssymb}
\usepackage{tabularx}
\usepackage{multirow}
\usepackage{booktabs}
\usepackage{subcaption}
\usepackage{caption}
\usepackage{soul}
\usepackage{afterpage}
\usepackage[dvipsnames]{xcolor}

\usepackage[pagebackref=true,breaklinks=true,letterpaper=true,colorlinks,bookmarks=false]{hyperref}

\newcommand{\Mat}{\boldsymbol}

\DeclareMathOperator{\mean}{\mathbb{E}}

\iccvfinalcopy %

\makeatletter
\let\ps@empty\ps@plain
\makeatother

\begin{document}

\title{Reference-based Painterly Inpainting via Diffusion: \\ Crossing the Wild Reference Domain Gap}

\author{%
  Dejia Xu\textsuperscript{1}, Xingqian Xu\textsuperscript{2,3}, Wenyan Cong\textsuperscript{1}, Humphrey Shi\textsuperscript{2, 3}, Zhangyang Wang\textsuperscript{1,3}\\
  {{\textsuperscript{1}University of Texas at Austin}, {\textsuperscript{2}SHI Labs @ Georgia Tech \& UIUC}, {\textsuperscript{3}Picsart AI Research (PAIR)}} \\
  \small{\texttt{\{dejia,atlaswang\}@utexas.edu}} \\
   \\
}

\twocolumn[{
\vspace{-3mm}
\renewcommand\twocolumn[1][]{#1}%
\maketitle
\begin{center}
    \vspace{-5mm}
    \includegraphics[width=1.0\textwidth]{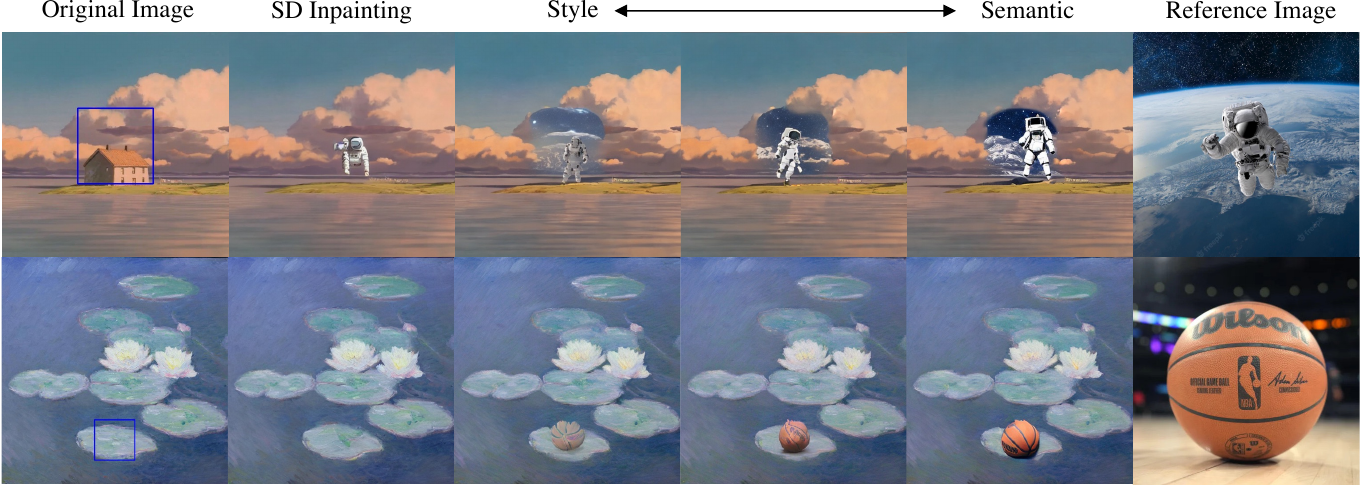}
    \captionof{figure}{
    Our proposed \underline{Ref}erence-based \underline{Paint}erly Inpainting framework (RefPaint) allows controlling the strength of reference semantics and background style when performing inpainting. Compared to Stable Diffusion, which uses text prompts as the reference\protect\footnotemark, our RefPaint captures the reference information better and is able to generate consistent styles.
    }
    \vspace{1mm}
    \label{fig:teaser}
\end{center}
}]

\begin{abstract}
Have you ever imagined how it would look if we placed new objects into paintings? 
For example, what would it look like if we placed a basketball into Claude Monet's ``Water Lilies, Evening Effect''?
We propose \underline{Ref}erence-based \underline{Paint}erly Inpainting, a novel task that crosses the wild reference domain gap and implants novel objects into artworks.
Although previous works have examined reference-based inpainting, they are not designed for large domain discrepancies between the target and the reference, such as inpainting an artistic image using a photorealistic reference.
This paper proposes a novel diffusion framework, dubbed RefPaint, to ``inpaint more wildly'' by taking such references with large domain gaps.
Built with an image-conditioned diffusion model, we introduce a ladder-side branch and a masked fusion mechanism to work with the inpainting mask. 
By decomposing the CLIP image embeddings at inference time, one can manipulate the strength of semantic and style information with ease.
Experiments demonstrate that our proposed RefPaint framework produces significantly better results than existing methods. Our method enables creative painterly image inpainting with reference objects that would otherwise be difficult to achieve.
Project page: \url{https://vita-group.github.io/RefPaint/}
\end{abstract}

\section{Introduction}

\footnotetext{The text prompts used for Stable Diffusion~\cite{rombach2022high} are ``an astronaut in front of the earth'' and ``a photo of a basketball''.}
Editing the contents within an image at will has always been an important task in computer vision. Human artists are especially good at manipulating images with their imaginations in mind. However, obtaining a visually pleasing image requires a lot of human effort. It has attracted researchers' interest for decades to relieve the burdens of humans by asking artificial intelligence for help. While numerous methods have been proposed, how to work with artistic images remains an open question.

In this paper, we present a novel task, named \textbf{\underline{Ref}erence-based \underline{Paint}erly Inpainting}, where we aim to inject new objects into an artistic image. Given a target image with masks, we also have access to a reference image used as context information. Our goal is to merge this novel object into the existing artwork while retaining the overall artistic style and deriving localized edits that resemble the reference image. Although similar to general reference-based inpainting, our task differs in the challenging domain gap between the reference object and the artistic background image.

\begin{table}[t]
\centering
\resizebox{\columnwidth}{!}{%
\begin{tabular}{l|cccc}
\toprule
\textbf{Task} & \textbf{Localized} & \textbf{Reference} & \textbf{Domain Gap} & \textbf{Diversity} \\
\midrule
Image Harmonization & \checkmark & $\times$ & $\times$ & $\times$ \\
Painterly Harmonization & \checkmark & $\times$ & \checkmark & $\times$  \\ 
Image Variation & $\times$  & $\times$  & $\times$  & \checkmark\\
Image Stylization & $\times$ & $\times$ &\checkmark & \checkmark \\
Image Inpainting & \checkmark & $\times$ & $\times$ & \checkmark \\
Text-based image inpainting & \checkmark & \checkmark & $\times$ & \checkmark \\
Reference-based image inpainting & \checkmark & \checkmark & $\times$ & $\times$ \\
\textbf{Reference-based Painterly Inpainting} & \checkmark & \checkmark & \checkmark & \checkmark \\ 
\bottomrule
\end{tabular}
}
\caption{Comparison between our proposed Reference-based Painterly Inpainting and several related tasks.}
\label{tab:tasks}
\end{table}

Image inpainting has been well-studied in the community for decades. Traditional approaches are based on hand-crafted heuristics~\cite{bertalmio2000image, barnes2009patchmatch}. Recent data-driven models utilize specialized network architecture~\cite{nazeri2019edgeconnect,pathak2016context,iizuka2017globally,liu2018image,yu2019free,suvorov2022resolution} or additional hint information~\cite{ren2019structureflow, liao2020guidance}.
Nevertheless, these blind image inpainting are not applicable when humans want to manipulate what to fill in. Consequently, reference-based image inpainting is introduced to borrow the available information from the reference. TransFill~\cite{zhou2021transfill} and GeoFill~\cite{zhao2022geofill} effectively learn the correspondence between the reference and input images. However, these methods do not take the painterly inpainting settings into consideration and are not suitable for the wild domain gap, where little correspondence can be found.

Another related line of work involves image harmonization\cite{pitie2005n,reinhard2001color,jia2006drag,perez2003poisson,tao2010error,zhu2015learning, cun2020improving, cong2020dovenet, jiang2021ssh}, where a given object is composited to a background. Painterly Harmonization~\cite{luan2018deep,zhang2020deep,peng2019element} specifically handles the case where the background is artistic, and style transfer is required. 
However, image harmonization usually directly pastes the exact object to the background, while our task requires generating semantically consistent objects that are not necessarily always identical.

The closest baseline to us is the Stable Diffusion~\cite{rombach2022high} Inpainting model implemented with RePaint~\cite{lugmayr2022repaint}
Recent success in text-to-image diffusion models enables humans to inpaint images using text prompts as reference. However, it is generally hard to provide more fine-grained references, such as images, when manipulating the contents. Although text-based image manipulation has gained tremendous interest, text is usually considered a vague specification and involves a great deal of ambiguity. One may argue that this is desirable since it might lead to diversity, but some may prefer to gain more precise control over the manipulated contents. Moreover, most approaches based on text-to-image diffusion models fail to capture redundant information in the reference image. Tab.~\ref{tab:tasks} compares our proposed Reference-based Painterly Inpainting with the above-mentioned similar tasks.

To this end, we propose a novel image-conditioned diffusion model that is capable of generating a plausible composition of the masked image and the inpainted object. We first implement an image-conditioned diffusion model to perform image variation (Versatile Diffusion~\cite{vd}), then adopt an additional ladder side branch and masked fusion block to introduce the mask control and the localized edits. We also perform principle component analysis (PCA) for the CLIP image embedding to decompose into semantics and style. By mixing the conditional generation of the classifier-free guidance, our model is able to perform fine-grained control of the trade-off between the reference semantics and background style.

Our main contributions can be summarized as follows,

\begin{itemize}
    \item We present a novel task, \underline{Ref}erence-based \underline{Paint}erly Inpainting, where we inpaint artistic images with an additional reference available from the real world. We propose a novel image-conditioned diffusion framework, dubbed RefPaint Diffusion, which can generate visually pleasing and coherent results despite the challenging domain gap between real-world reference objects and artistic background images.
    \item We introduce a ladder-side branch into the Versatile Diffusion~\cite{vd} backbone and a masked fusion mechanism to inject additional mask information for localized image manipulations. By decomposing the conditioning CLIP image embeddings at inference time, we are able to control to what extent we want the final results to lean toward semantic alignment with real-world reference objects or style alignment with artistic background images.
    \item We conduct extensive experiments using artistic images from the WikiArt dataset~\cite{saleh2015large} and real-world objects from COCO Captions dataset~\cite{chen2015microsoft}. The results demonstrate that our proposed RefPaint Diffusion outperforms existing approaches in terms of diversity and fidelity, both qualitatively and quantitatively.
\end{itemize}

\section{Related Work}

\subsection{Image Inpainting}

Traditional image inpainting models are mostly built using hand-crafted heuristics.
Diffusion-based methods~\cite{bertalmio2000image} propagate pixel colors from the background regions to the masked holes and are limited to small hole sizes and images with fewer texture variations. Alternatively, patch-based approaches~\cite{wexler2004space,barnes2009patchmatch} search for similar regions in the image to complete missing patches. While these approaches can provide high-quality details by copy-pasting patches, the filled regions are usually inconsistent with the surrounding regions due to the lack of a high-level global structural understanding of the image. 
Recently, data-driven models have prevailed thanks to the outstanding abilities of neural networks.
Context encoder~\cite{pathak2016context} learns to encode the surrounding regions. Iizuka~et~al.~\cite{iizuka2017globally} adopt multiple discriminators to ensure local and global consistency for the final result. Special convolution operations are also proposed to work with the inpainting mask, such as partial convolution~\cite{liu2018image},  gated convolution~\cite{yu2019free},  and Fourier convolution~\cite{suvorov2022resolution}. Despite great advances in blind inpainting, it remains a highly ill-posed problem. Researchers have discovered that additional information can benefit the inpainting quality greatly, such as edges~\cite{nazeri2019edgeconnect}, segmentation masks~\cite{song2018spg}, low-frequency structures~\cite{ren2019structureflow,liao2020guidance} and stereo or multi-view images~\cite{wang2008stereoscopic,bhavsar2010inpainting,baek2016multiview,ma2020learning,ma2021fov,thonat2016multi}. 
The line of work more relevant to ours is reference-based inpainting, where an additional reference image is available for the inpainting task. 
TransFill~\cite{zhou2021transfill} performs reference-guided inpainting by warping a reference image via homography. 
GeoFill~\cite{zhao2022geofill} leverages monocular depth maps, and relative camera pose to understand the 3D geometry.
However, existing works are mainly designed for borrowing pixel-level information from the reference image, and thus are not applicable to the problem of painterly inpainting, where we need to cross the large domain gap between the background image and the reference image.

\begin{figure*}
\centering
\includegraphics[width=\textwidth]{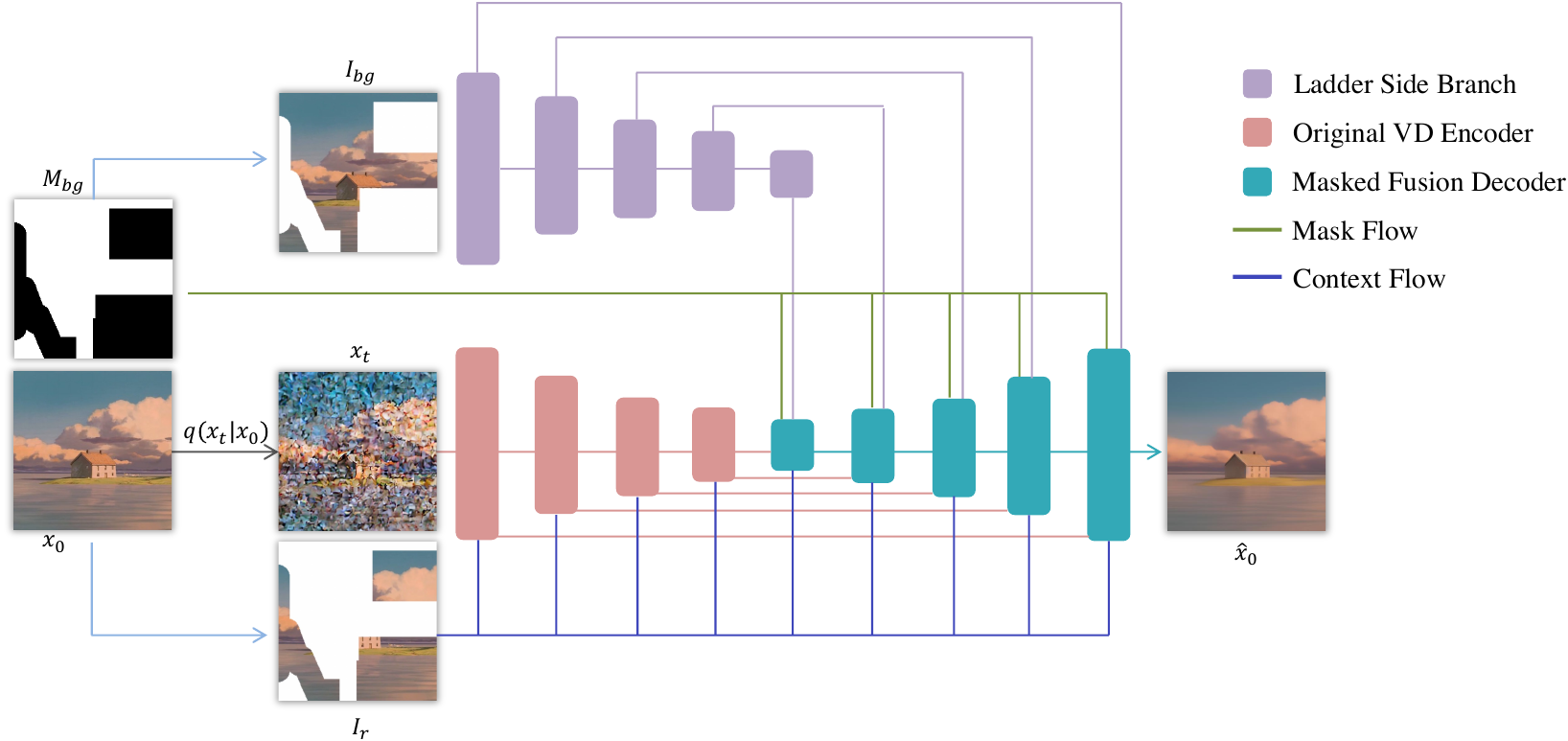}
\caption{An overview of our proposed RefPaint framework. We utilize a ladder-side branch and a masked fusion block to incorporate additional mask information. The framework is trained in a self-supervised manner.}
\vspace{-2mm}
\label{fig:framework}
\end{figure*}

\subsection{Diffusion Model}

Denoising diffusion model~\cite{sohl2015deep} is a type of deep generative model that synthesize data through an iterative denoising process. Diffusion models consist of a forward process that adds
noise to clean images and a reverse process that learns to denoise. They have demonstrated outstanding image generation~\cite{song2019generative,ho2020denoising} capability with the help of various improvements in architecture design~\cite{dhariwal2021diffusion}, sampling guidance~\cite{ho2022classifier}, and inference cost~\cite{salimans2022progressive, rombach2022high, vahdat2021score}. Equipped with large-scale image-pair datasets, many works scaled up the model ~\cite{rombach2022high, nichol2021glide, ramesh2022hierarchical, saharia2022photorealistic} to billions of parameters to work with the challenging text-to-image generation. Among them, Stable Diffusion~\cite{rombach2022high} gained tremendous popularity among the community, which reduced the computation cost by applying the diffusion process to the low-resolution latent space instead of directly in the pixel space. Additionally, denoising diffusion models have also found success in a number of various computer vision tasks, including text-to-3D~\cite{poole2022dreamfusion,singer2023text}, image-to-3D~\cite{xu2022neurallift}, 3D reconstruction~\cite{gu2023nerfdiff,watson2022novel}, object detection~\cite{chen2022diffusiondet}, depth estimation~\cite{saxena2023monocular}, etc.
Versatile Diffusion~\cite{vd} studies to unify multiple workflows of text and image generation into one multimodal model. Such a design initiates novel extensions, such as image variation, with the diffusion model. In this work, we build our framework based on versatile diffusion~\cite{vd}, since it is an open-source image-conditioned diffusion model.

\subsection{Image Harmonization}

Given a composite image, image harmonization aims to adjust the foreground appearance to make it compatible with the background.
Traditional image harmonization methods focus mainly on adjusting low-level appearance statistics~\cite{pitie2005n,reinhard2001color,jia2006drag,perez2003poisson,tao2010error}. More recently, learning-based methods attempted to learn a neural network for harmonization~\cite{zhu2015learning, cun2020improving, cong2020dovenet, jiang2021ssh}.
Painterly image harmonization aims to transfer the background style to the foreground while retaining the foreground content. 
Luan \textit{et al.}~\cite{luan2018deep} determines the local statistics be transferred by ensuring both spatial and inter-scale statistical consistency. Zhang \textit{et al.}~\cite{zhang2020deep} jointly optimized a novel Poisson gradient loss with content and style loss. Peng \textit{et al.}~\cite{peng2019element} adopts AdaIN to manipulate the style and global and local discriminators for adversarial learning.
Note that for image harmonization, the exact object is directly pasted to the background, while our task is to generate semantically consistent objects but not necessarily to be always identical.

\section{RefPaint Diffusion}

We define the \underline{Ref}erence-based \underline{Paint}erly Inpainting problem as: Given input quadruplets ($I_r, I_{bg}, M_o, M_{bg}$) that consists of an object-centric reference image $I_r \in R^{H_r \times W_r \times 3}$, a background image $I_{bg} \in R^{H_{bg} \times W_{bg} \times 3}$ and their corresponding binary masks $M_o \in R^{H_o \times W_r \times 1}$ and $M_{bg} \in R^{H_{bg} \times W_{bg} \times 3}$, the goal is to inpaint the input object $I_r$ into the masked region $I_{bg} \times M_{bg}$. For the object-centric binary mask $M_r$, the desired object location is set to 1, and the rest is set to 0. For the background mask $M_{bg}$, the desired location for inpainting is set to 0, and the rest regions asked to remain untouched are set to 1. The overall output result should be visually coherent and consistent with the masked background regions $M_{bg}$ remaining untouched, and the inpainted objects preserving the semantics of the reference object $I_r$. Our problem setting differs from text-guided image inpainting in that our reference information is an object-centric image rather than a text prompt. Our problem is also different from image harmonization, where the generated composite image should preserve the semantics instead of the exact appearance of the original object.

An overview of our proposed framework is provided in Fig.~\ref{fig:framework}. To leverage the outstanding generation ability of pre-trained diffusion models, we build our RefPaint Diffusion based on a pre-trained image-conditioned diffusion model. We construct a ladder-side finetuning technique as well as a masked fusion mechanism to introduce our inpainting masks into the diffusion model.
Our model is trained in a self-supervised manner without the need to curate any annotations.
During inference time, we decompose the CLIP image embeddings via PCA and perform disentangled classifier-free guidance to manipulate the strength of the reference semantic and the background style.

\subsection{Preliminary of Diffusion Model}

Diffusion models learn to generate images via iterative refinement. 
The forward diffusion process $q(\Mat{x}_{t} | \Mat{x}_{t-1})$ involves adding Gaussian noise to clean images $\Mat{x}_0$. 
With curated $\alpha_t$ and $\sigma_t$, their marginal distribution can be derived as $q\left(\Mat{x}_t \mid \Mat{x_0}\right)=\mathcal{N}\left(\alpha_t \Mat{x_0}, \sigma_t^2 \Mat{I}\right)$, and converges to $\mathcal{N}(\Mat{0}, \Mat{I})$ when $t$ reaches the end of the forward process~\cite{kingma2021variational,song2020score}. The reverse diffusion process $p(\Mat{x}_{t-1} | \Mat{x}_t)$ is approximated by the Gaussian distribution~\cite{sohl2015deep,song2020score} and is formulated by a denoising network $\Mat{\epsilon}_\theta(\Mat{x}_t, t)$.

\looseness=-1
The diffusion models $\Mat{\epsilon}_\theta(\Mat{x}_t, t)$ take noisy images as inputs and estimate the noise components at each timestep. They are trained via optimizing the weighted evidence lower bound (ELBO)~\cite{ho2020denoising,kingma2021variational}:
\begin{equation}
\mathcal{L}_{\text{ELBO}}(\theta)= \mean\left[w(t)\left\|\Mat{\epsilon}_\theta\left(\alpha_t \Mat{x}_0+\sigma_t \Mat{\epsilon} ; t\right)-\Mat{\epsilon}\right\|_2^2\right],
\label{eq:ddpm}
\end{equation}
where $\Mat{\epsilon} \sim \mathcal{N}(\mathbf{0}, \mathbf{I})$. $w(t)$ is a weighting function found to perform well when set to $w(t)=1$~\cite{ho2020denoising}.
There are stochastic \cite{ho2020denoising}, and deterministic \cite{song2020denoising} approach to sample from a diffusion model. After sampling $\Mat{x}_T \sim \mathcal{N}(\mathbf{0}, \mathbf{I})$, we iteratively reduce the noise level and gradually reach a clean image.

Text-to-image diffusion models condition the noise prediction on the text prompt using cross-attention ~\cite{chen2021crossvit} layers. Versatile Diffusion~\cite{vd} extends such conditioning information to image embeddings to support image variation and image-to-text generation. 
Although we built on Versatile Diffusion~\cite{vd} in our experiments, our proposed framework is independent of the choice of diffusion model backbone and should be able to work with other implementations of image-conditioned diffusion models.

\subsection{RefPaint Modules}

We provide an overview of our proposed framework in Fig.~\ref{fig:framework}. We build on the Versatile Diffusion~\cite{vd} backbone and add an additional ladder side branch to take the masked image as input. In order to ensure that information from both encoding branches get fully utilized, we adopt a masked fusion block to blend the information using the inpainting mask. In the next few paragraphs, we will illustrate these designs in detail.

\paragraph{Ladder Side Branch}

We design a ladder-side finetuning branch to work with the additional mask information. More specifically, we don't change the original workflow of the Versatile Diffusion~\cite{vd} (VD), but instead, we include a new encoding branch with the same architecture onto the vanilla UNet. The new ladder-side branch doesn't take VAE-encoded latent as in the original VD architecture, but instead directly operates on images, thus obtaining more fine-grained conditioning information.
In order to make the network aware of the inpainting masks, we fine-tune the model using masked CLIP image embeddings.

Additionally, we feed the reference object $I_r$ through the context flow of the vanilla VD UNet, and feed the image to be inpainted $I_{bg}$ through the ladder-side branch. This design stabilizes the fine-tuning process since it aligns with the purpose of the VD checkpoint originally pre-trained for the image variation task.
Note that although ControlNet~\cite{zhang2023adding} adopted a similar approach, their design does not support masks or handle localized edits of the images.

\begin{figure}
    \centering
    \includegraphics[width=0.9\columnwidth]{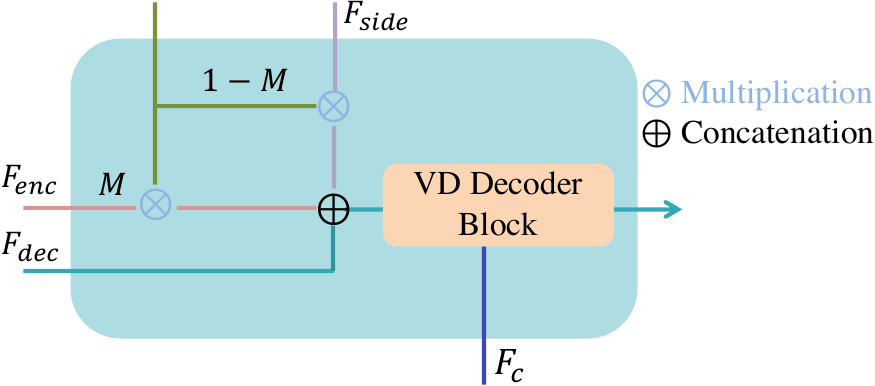}
    \caption{Structure of our masked fusion block. We inject a masked fusion mechanism in front of the original VD decoder block, to work with the features extracted from the ladder side branch.}
    \label{fig:block}
\end{figure}

\paragraph{Masked Fusion Block}

A detailed workflow of the masked fusion block is provided in Fig.~\ref{fig:block}.
More specifically, we use the mask $M$ to adaptively blend the features maps, formulated as follows:
$$
F = \text{cat}[F_\text{side} \times (1 - M) + F_\text{enc} \times M, F_\text{dec}],
$$
where $\text{cat}$ denotes concatenation operation and $F$ is the input to the original VD Decoder block. Such a design does not require changing the VD backbone architecture. Note that for simplicity, we omit the downsampling operation on the mask $M$.

We hypothesize that the intermediate feature maps after each block in the UNet are spatially associated with the corresponding input. For example, the features on the top left corner will be related to the top left input image, while the features on the bottom right will represent mostly the bottom right of the image.
To efficiently utilize the redundant information from the two branches and ensure the network adapts to additional masks, we perform masked blending of the two feature streams instead of directly fusing them. Since inpainting masks are not explicitly fed into the diffusion network, this masked fusion mechanism helps the two branches learn consistent features and prevents the network from simply ignoring the features from either branch.

\paragraph{Self-supervised Training}
The training of our framework requires ground truth for the desired image composition. However, such ground truths are hard to obtain for in the wild image pairs since they require great human expertise. To overcome this issue, we propose to decompose a single image into two regions instead of seeking for some well-annotated compositions. Specifically, for a given image in the dataset $I$ and a random binary mask $M$, we can construct the quadruplets as follows:
$$
\begin{aligned}
I_{bg} & = I \otimes M, \\
I_{o} &= I \otimes (1 - M), \\
M_{o} &= M, \\
M_{bg} &= 1 - M.
\end{aligned}
$$
During training, $I$ is the $x_0$ in Fig.~\ref{fig:framework}. We first add noise to obtain $x_t$ and then denoise by the diffusion model to optimize the ELBO as in Eq.~\ref{eq:ddpm}. At inference time, we start the reverse diffusion process from Gaussian noise and gradually reduce the noise level.
Our masked fusion mechanism prevents the model from overfitting on direct stitching of the two branches.

\paragraph{Disentangled Semantic and Style Fusion via Classifier-free Guidance}
For our reference-based painterly inpainting task, we aim to mix the style information from the background image with the semantic information from the reference object. We utilize classifier-free guidance~\cite{ho2022classifier} to overcome the challenging domain gap between the background and the reference.
Classifier-free guidance~\cite{ho2022classifier} is one of the most widely used techniques for conditional diffusion models to improve sampling quality. 
We jointly learn a conditional diffusion model and an unconditional diffusion model by randomly dropping the conditional information during some training iterations. 
During inference time, we amplify the conditional likelihood and suppress the unconditional likelihood. At each timestep, we inference both models and obtain a weighted combination of the results.

\looseness=-1
Specifically, in our framework, we further decompose the CLIP image embeddings using PCA. The low-rank components correspond to the semantics $c_\text{sem}$, while the rest mainly represent the style of the image $c_\text{sty}$.
$$
\begin{aligned}
\tilde\epsilon_{\theta}(x_t, c_\text{ref}, c_\text{bg})  &= (1 - \omega) \epsilon_{\theta}(x_t, \phi) + \omega\gamma\epsilon_{\theta}(x_t, c_\text{ref, sem})\\
&  + \omega(1 - \gamma)\epsilon_{\theta}(x_t, c_\text{bg, sty})),
\end{aligned}
$$
where $\omega$ and $\gamma$ are weighting coefficients that control the strength of the conditional likelihood and the trade-off between the reference semantic and the background style.
Larger $\gamma$ leads to better semantic alignment, while smaller leads to better style alignment.
As a result, we inherently support controlling the strength of semantics and style during inference time by tuning these weights of classifier-free guidance.

\subsection{Training Recipe Tailored for Inpainting}

\paragraph{Token Masking Mechanism}
Since we want to calculate the CLIP image embedding of local regions inside images, we need to implement a masking mechanism to prevent the CLIP image model from seeing the masked-out regions. While multiplying the images directly with binary masks can be a straightforward approach, we argue that this will lead the model to learn from CLIP image embeddings that represent large regions of black regions.
We empirically found that directly masking the images before feeding them into the CLIP image encoder is suboptimal. This is partially due to the fact that such an operation will largely shift the image distribution away from that of the pre-trained image-conditioned diffusion model and make the model lean towards generation with black regions.

\paragraph{Mask Generation Strategy}
Since our self-supervised training require decomposing our training images using binary masks $M_{bg}$, we follow approaches similar to previous free-form inpanting work~\cite{yu2019free}.
The free-form masks consist of eraser-like holes that are simulated by randomly drawing lines and rotating angles repeatedly. To further ensure the smoothness of the generated masks, we place circle holes at the joints of the masks.
We additionally randomly set the masks in 25\% iterations to be full 0, which indicates that all the background pixels are masked out, and the model is asked to perform image variation using the new context branch.
Although object-centric masks are supposed to provide better image generation ability, we only experiment with random masks for simplicity and defer such extension to future works.

\paragraph{Color Constancy Mechanism}
During inference time, we observe that sometimes color drift appears for the image variation model. We suppose that this might be due to the fact that the backbone model was trained on reconstruction MSE loss and is not asked to preserve the white balance well.
To overcome this issue, we adopt a blending technique similar to RePaint~\cite{lugmayr2022repaint}. During the denoising diffusion process, we also prepare noisy versions of the background image $I_{bg, t} = \alpha I_{bg} + \sigma \epsilon_{bg}$, where $t$ is the corresponding time step. 
For early time steps where the noise level is high, we blend the noisy image $x_t$ with the original background images using the mask available: $x_t = x_t \otimes (1 - M_{bg}) + I_{bg, t} \otimes M_{bg}$.

\section{Experiment}

\subsection{Implementation Details}
We implement our image-conditioned diffusion model using Versatile Diffusion~\cite{vd}. The training is carried out self-supervised for 9,000 iterations. In each iteration, we use a batch size of 64 per GPU and 8 GPUs in total. 
The learning rate is $1e-4$, and we accumulate the gradient every four iterations. 
Before training, we initialize the ladder-side branch and original VD backbone using a pre-trained VD checkpoint for image variation.
During inference, $\omega$ is set to 7.5 as in previous diffusion models. 

\subsection{Dataset}
The training of RefPaint Diffusion is performed on the LAION-2b dataset~\cite{schuhmann2022laion}. Since we adopt a self-supervised training mechanism, we do not require any human annotations and can work with any dataset.
For testing, we randomly construct data pairs from the following two datasets.
Our background images are constructed using a random subset of 10k images from the Wikiart dataset~\cite{saleh2015large}. We first resize the images so that the shortest edge is $512$px and then perform a center crop to obtain the background $512 \times 512$ images.
Our object-centric images are randomly sampled from the COCO Captions dataset~\cite{chen2015microsoft}. We use ground truth segmentation annotations to locate the objects.
Then we randomly crop a $512 \times 512$ bounding box covering the largest object inside the image. 

\subsection{Baseline Methods}

The simplest baseline for our task is to ``Copy and Paste'', which directly paste the reference image onto the background. 
We compare against text-based inpainting using the Stable Diffusion model ``runwayml/stable-diffusion-inpainting''~\cite{rombach2022high}. For text prompts, we use a pre-trained BLIP model~\cite{li2022blip} to obtain the corresponding text captions from images and objects. We also include LaMa~\cite{suvorov2022resolution} in our experiments as a blind image inpainting method.

\subsection{Quantitative Comparisons}
In this section, we provide quantitative comparisons against existing methods.
We first show that the results from SD inpainting~\cite{rombach2022high} are very close to the results from LaMa~\cite{suvorov2022resolution}, indicating that SD inpainting usually ignores the reference text that describes the object and resembles blind inpainting. 
As there is no ground truth for our Reference-based Painterly Inpainting, we can not use pixel-wise full-reference metrics such as PSNR.
Instead, we calculate the CLIP image distance between the output inpainted image and the original artwork. We observe in Tab.~\ref{tab:baselines} that SD Inpainting generates similar numbers with LaMa. Our method, on the other hand, successfully inpaints the object into the background image and reaches a larger CLIP image distance. A larger CLIP image distance implies that our results are not simply painting the background like SD inpainting, but include the reference object.
Additionally, we measure the CLIP image distance between the object regions of the output inpainted images and the CP baseline's results. 
This metric can reveal how much the object is being inpainted. Numbers are calculated on the 10k images subset we randomly sampled.

{
\begin{figure*}
    \centering
\begin{subfigure}{0.19 \textwidth}
\includegraphics[width=\textwidth]{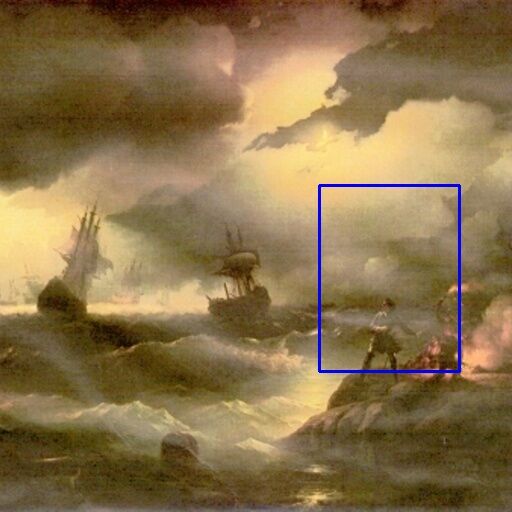}
\end{subfigure}
\begin{subfigure}{0.19\textwidth}
\includegraphics[width=\textwidth]{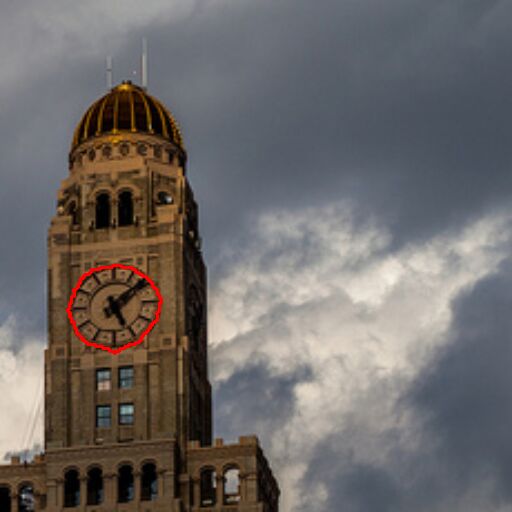}
\end{subfigure}
\begin{subfigure}{0.19 \textwidth}
\includegraphics[width=\textwidth]{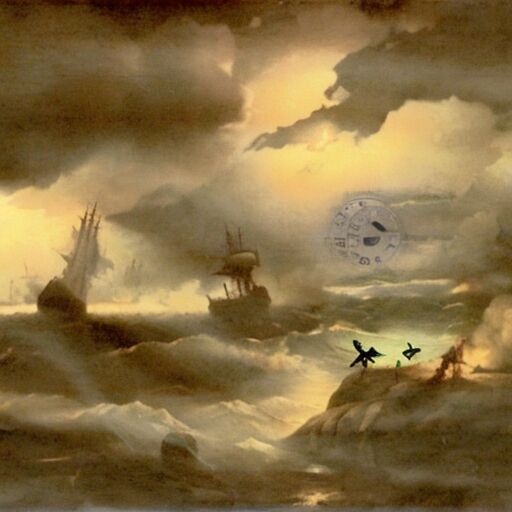}
\end{subfigure}
\begin{subfigure}{0.19 \textwidth}
\includegraphics[width=\textwidth]{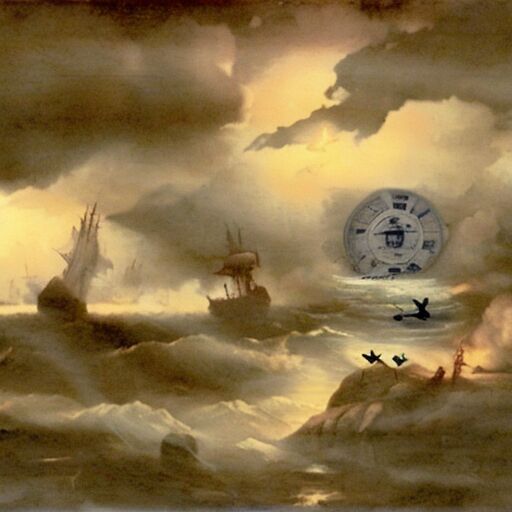}
\end{subfigure}
\begin{subfigure}{0.19 \textwidth}
\includegraphics[width=\textwidth]{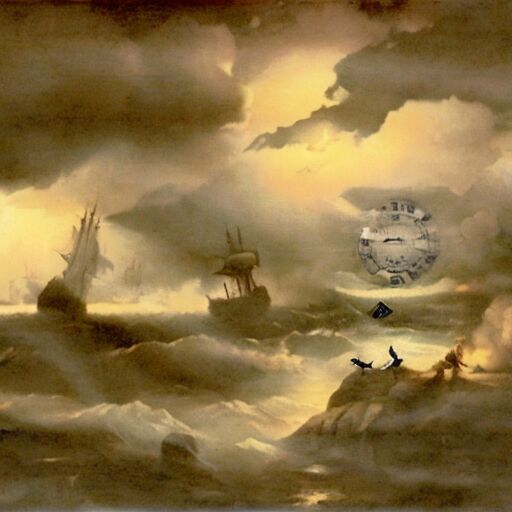}
\end{subfigure}
\\
\begin{subfigure}{0.19 \textwidth}
\includegraphics[width=\textwidth]{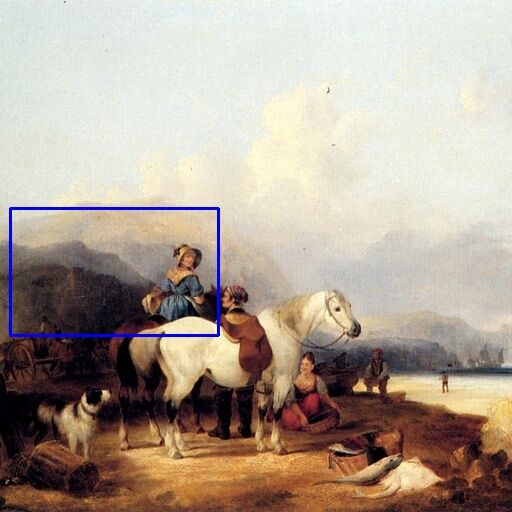}
\end{subfigure}
\begin{subfigure}{0.19\textwidth}
\includegraphics[width=\textwidth]{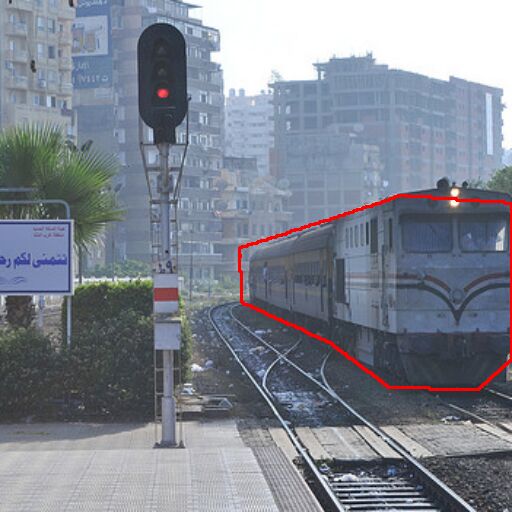}
\end{subfigure}
\begin{subfigure}{0.19 \textwidth}
\includegraphics[width=\textwidth]{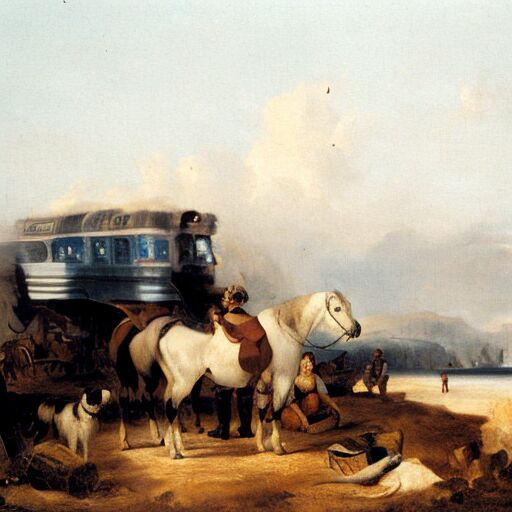}
\end{subfigure}
\begin{subfigure}{0.19 \textwidth}
\includegraphics[width=\textwidth]{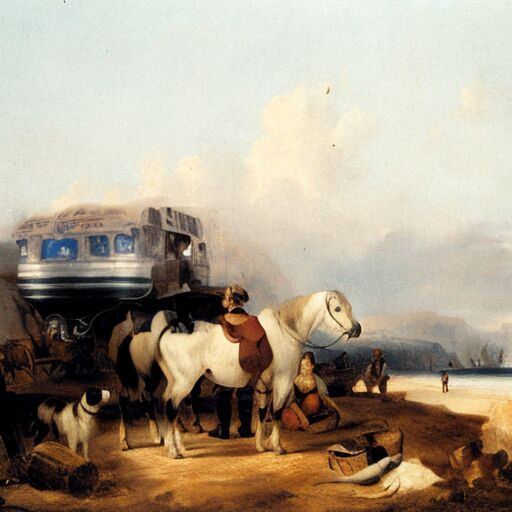}
\end{subfigure}
\begin{subfigure}{0.19 \textwidth}
\includegraphics[width=\textwidth]{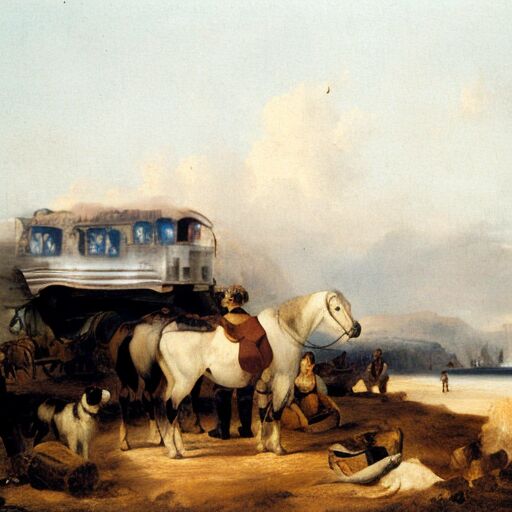}
\end{subfigure}
\\
\begin{subfigure}{0.19 \textwidth}
\includegraphics[width=\textwidth]{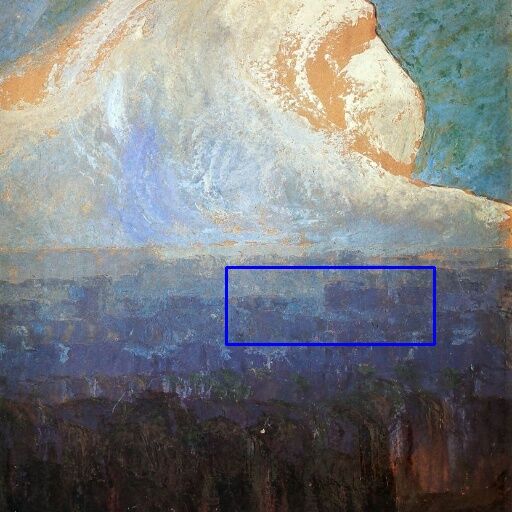}
\end{subfigure}
\begin{subfigure}{0.19\textwidth}
\includegraphics[width=\textwidth]{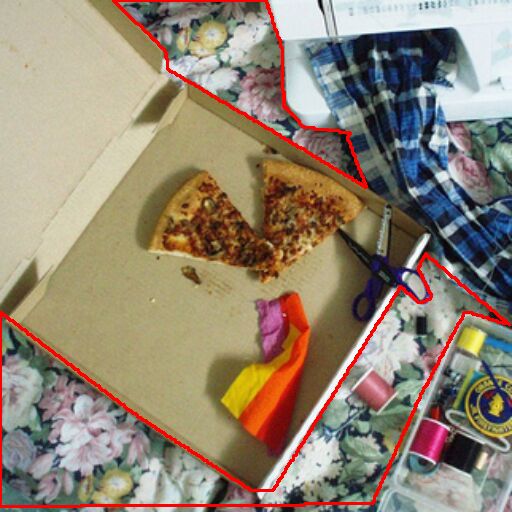}
\end{subfigure}
\begin{subfigure}{0.19 \textwidth}
\includegraphics[width=\textwidth]{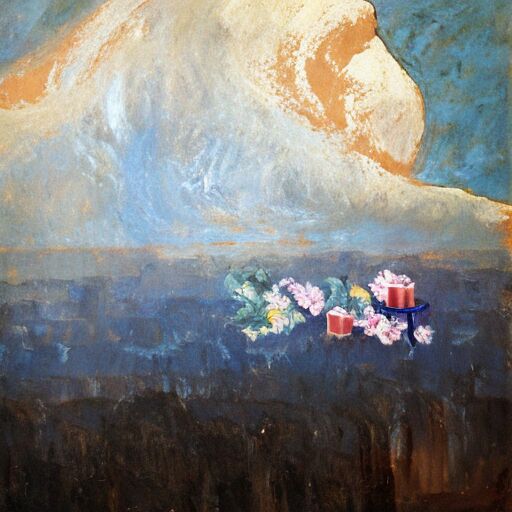}
\end{subfigure}
\begin{subfigure}{0.19 \textwidth}
\includegraphics[width=\textwidth]{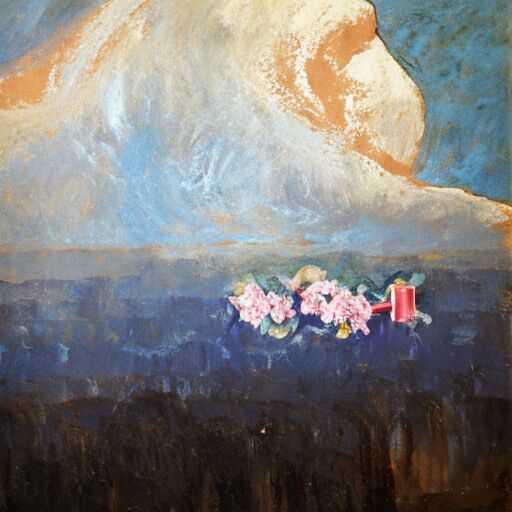}
\end{subfigure}
\begin{subfigure}{0.19 \textwidth}
\includegraphics[width=\textwidth]{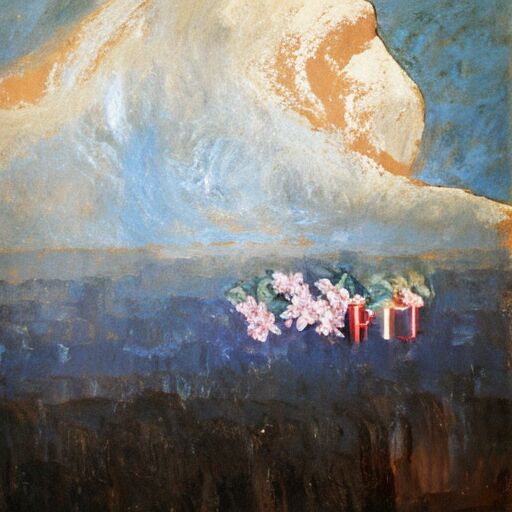}
\end{subfigure}
\\
\begin{subfigure}{0.19 \textwidth}
\includegraphics[width=\textwidth]{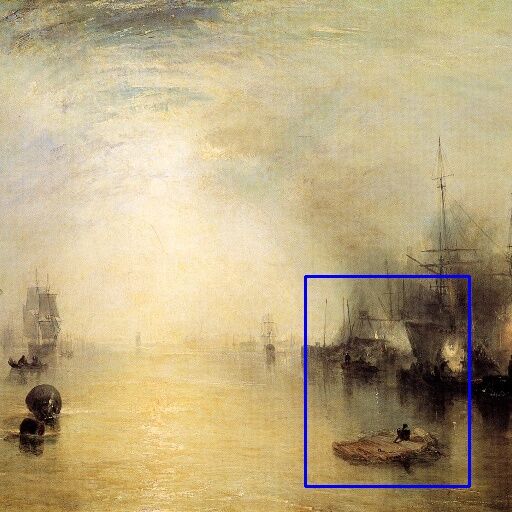}
\end{subfigure}
\begin{subfigure}{0.19\textwidth}
\includegraphics[width=\textwidth]{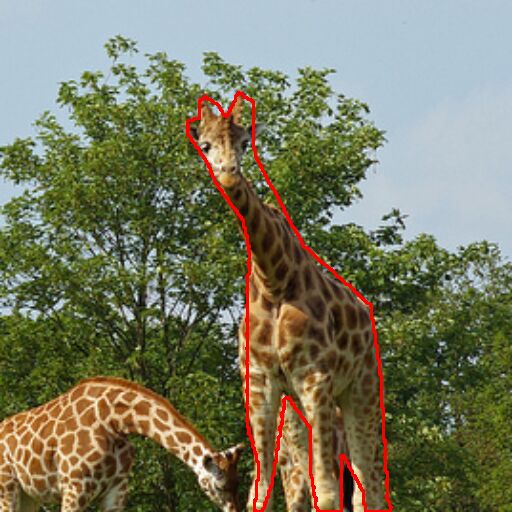}
\end{subfigure}
\begin{subfigure}{0.19 \textwidth}
\includegraphics[width=\textwidth]{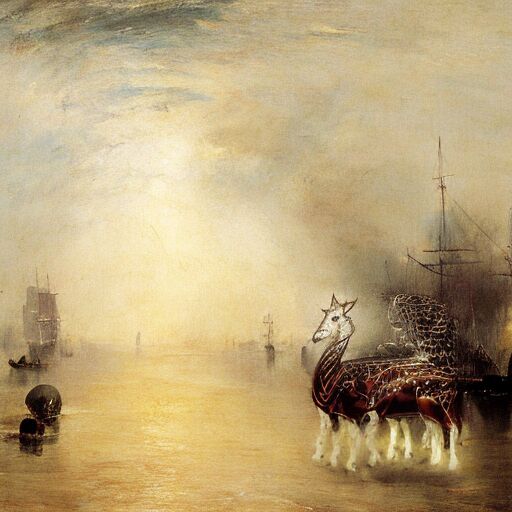}
\end{subfigure}
\begin{subfigure}{0.19 \textwidth}
\includegraphics[width=\textwidth]{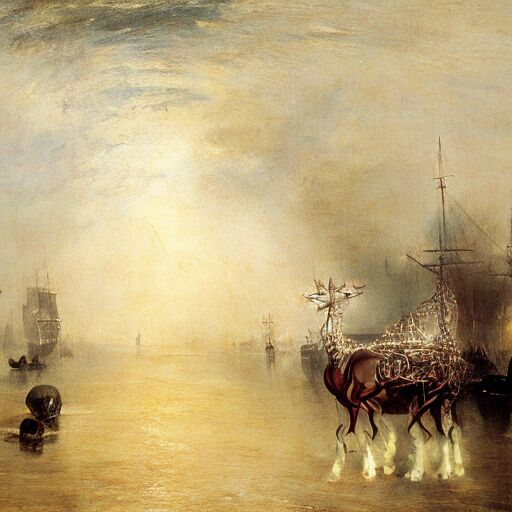}
\end{subfigure}
\begin{subfigure}{0.19 \textwidth}
\includegraphics[width=\textwidth]{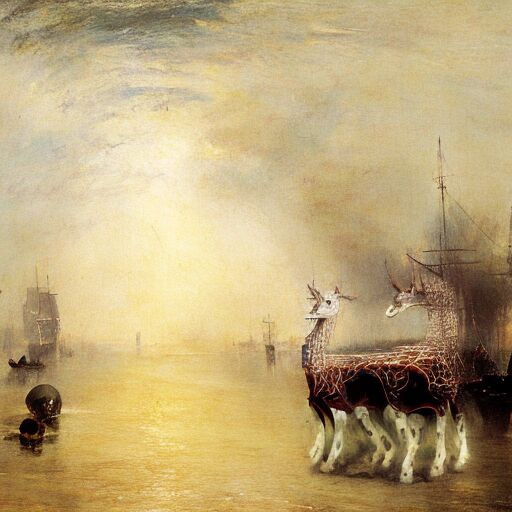}
\end{subfigure}
\\
\begin{subfigure}{0.19 \textwidth}
\includegraphics[width=\textwidth]{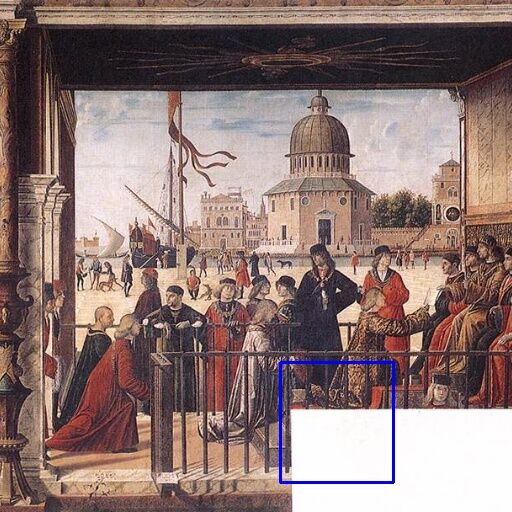}
\end{subfigure}
\begin{subfigure}{0.19\textwidth}
\includegraphics[width=\textwidth]{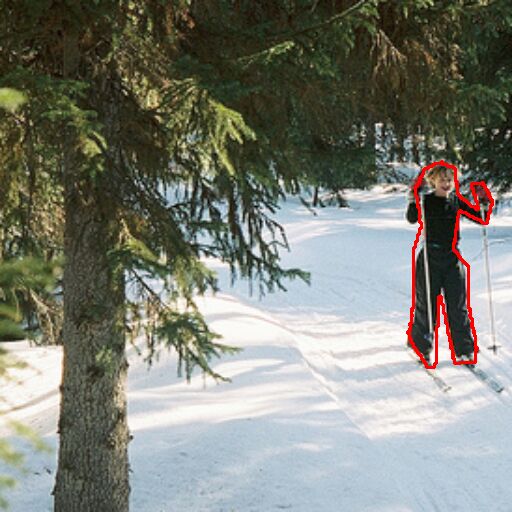}
\end{subfigure}
\begin{subfigure}{0.19 \textwidth}
\includegraphics[width=\textwidth]{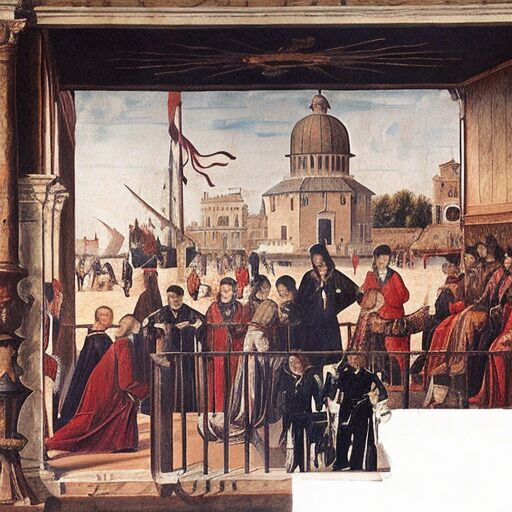}
\end{subfigure}
\begin{subfigure}{0.19 \textwidth}
\includegraphics[width=\textwidth]{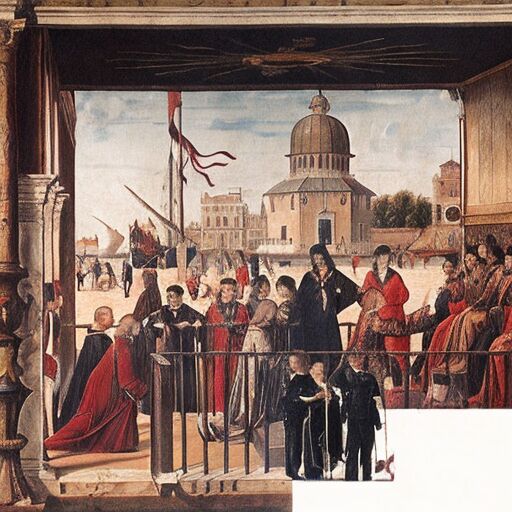}
\end{subfigure}
\begin{subfigure}{0.19 \textwidth}
\includegraphics[width=\textwidth]{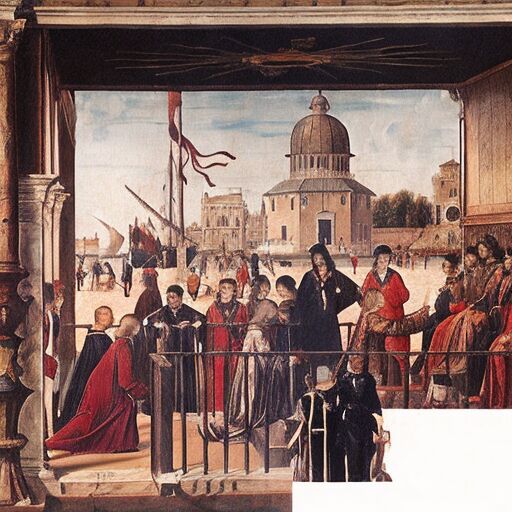}
\end{subfigure}
\\
\begin{subfigure}{0.19 \textwidth}
\includegraphics[width=\textwidth]{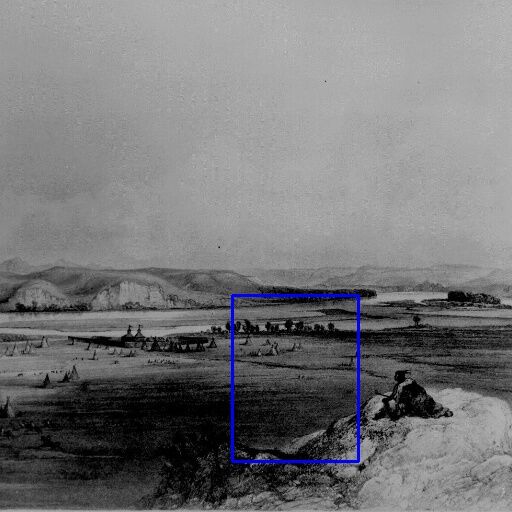}
\end{subfigure}
\begin{subfigure}{0.19\textwidth}
\includegraphics[width=\textwidth]{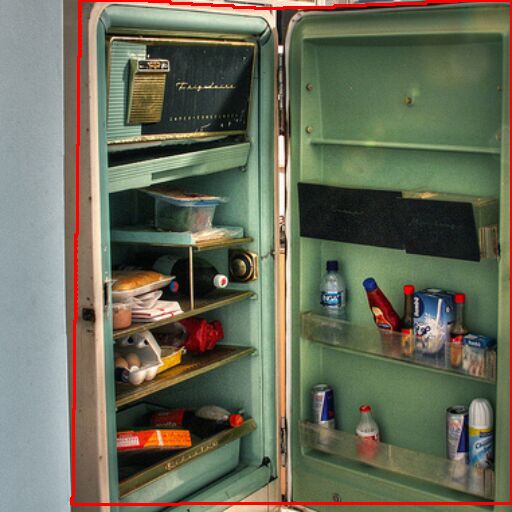}
\end{subfigure}
\begin{subfigure}{0.19 \textwidth}
\includegraphics[width=\textwidth]{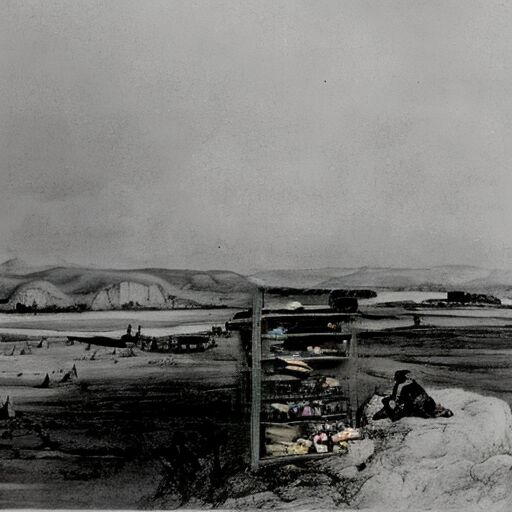}
\end{subfigure}
\begin{subfigure}{0.19 \textwidth}
\includegraphics[width=\textwidth]{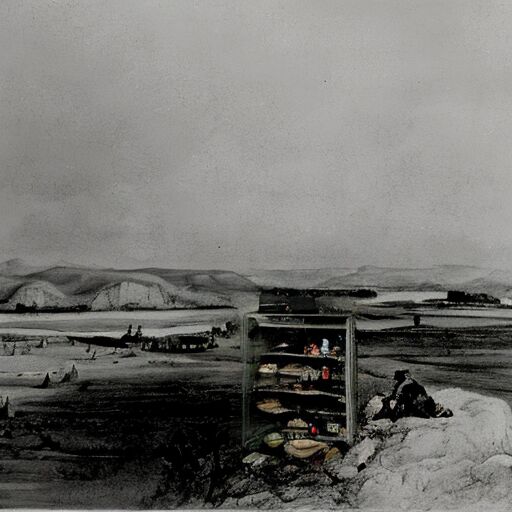}
\end{subfigure}
\begin{subfigure}{0.19 \textwidth}
\includegraphics[width=\textwidth]{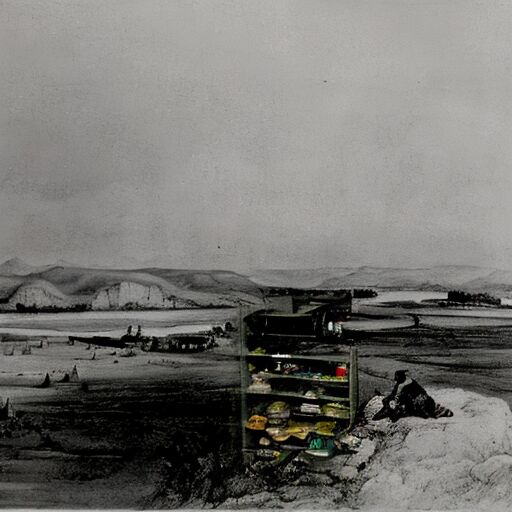}
\end{subfigure}
    \caption{Visual results for \underline{Ref}erence-based \underline{Paint}erly Inpainting results using \textbf{random} inpainting masks and \textbf{random} objects  from COCO Captions dataset. Blue bounding box represents the edited regions where we would like to inpaint. Red boundaries indicate the reference object. We provide \textbf{more comparisons} in supplementary.}
    \label{fig:results}
\end{figure*}
}

\begin{figure}
    \centering
    \begin{subfigure}{0.32 \columnwidth}
        \includegraphics[width=\textwidth]{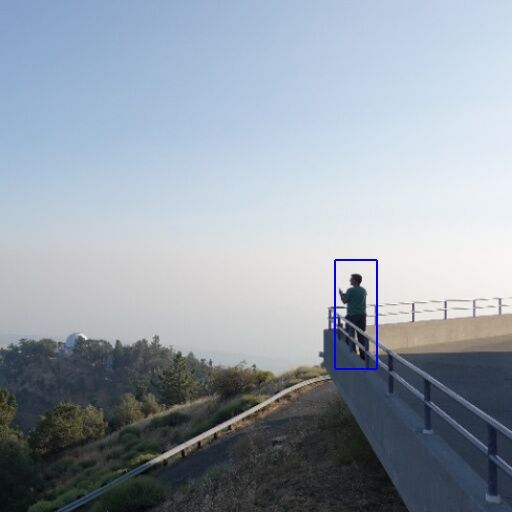}
    \end{subfigure}
    \begin{subfigure}{0.32 \columnwidth}
        \includegraphics[width=\textwidth]{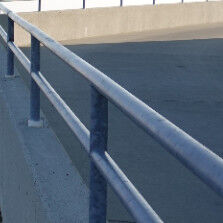}
    \end{subfigure}
    \begin{subfigure}{0.32 \columnwidth}
        \includegraphics[width=\textwidth]{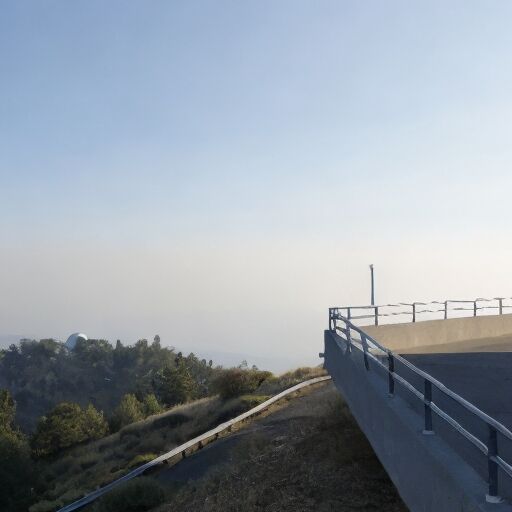}
    \end{subfigure}
    \\
    \begin{subfigure}{0.32 \columnwidth}
        \includegraphics[width=\textwidth]{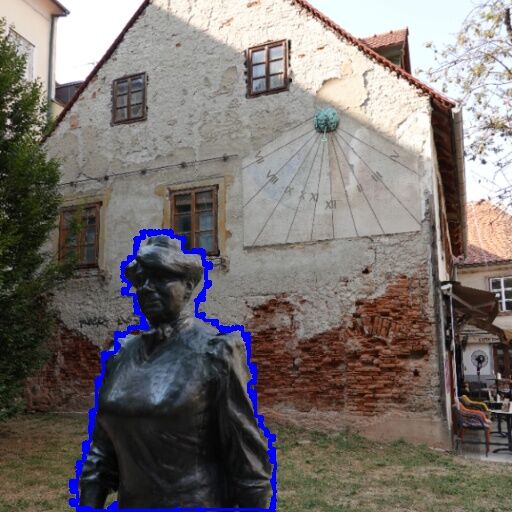}
        \caption{Original image}
    \end{subfigure}
    \begin{subfigure}{0.32 \columnwidth}
        \includegraphics[width=\textwidth]{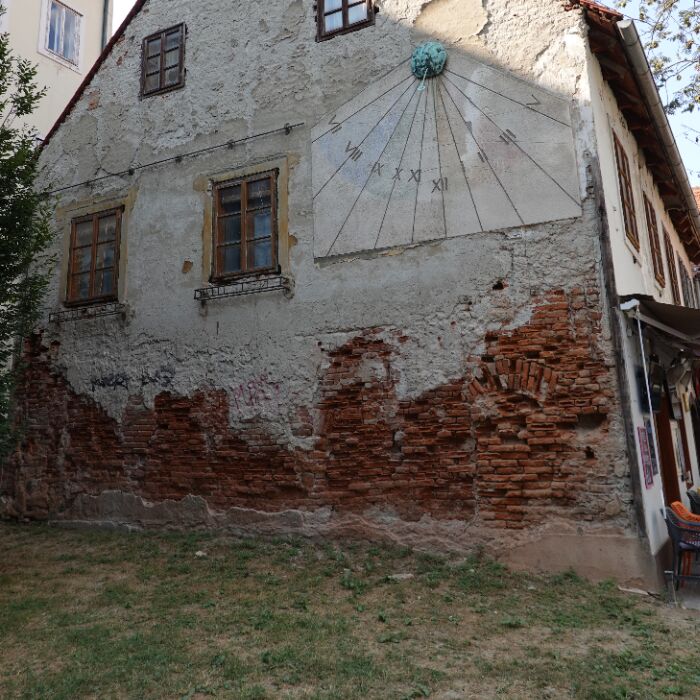}
        \caption{Reference}
    \end{subfigure}
    \begin{subfigure}{0.32 \columnwidth}
        \includegraphics[width=\textwidth]{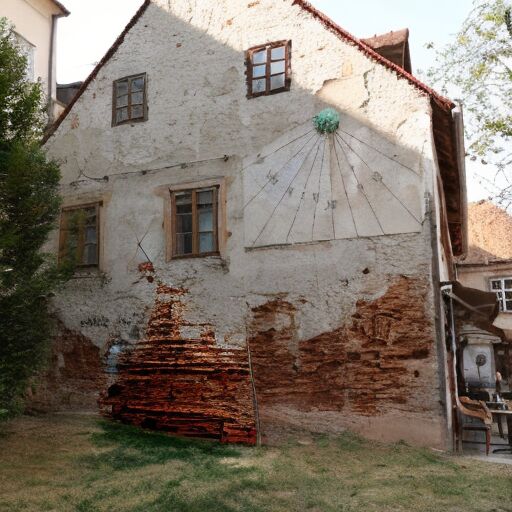}
        \caption{Output}
    \end{subfigure}
    \caption{Visual results of our RefPaint Diffusion when applied to the general reference-based inpainting. The blue bounding box indicates the inpainting mask.}
    \label{fig:transfill}
\end{figure}

\begin{figure}
    \centering
    \begin{subfigure}{0.32 \columnwidth}
        \includegraphics[width=\textwidth]{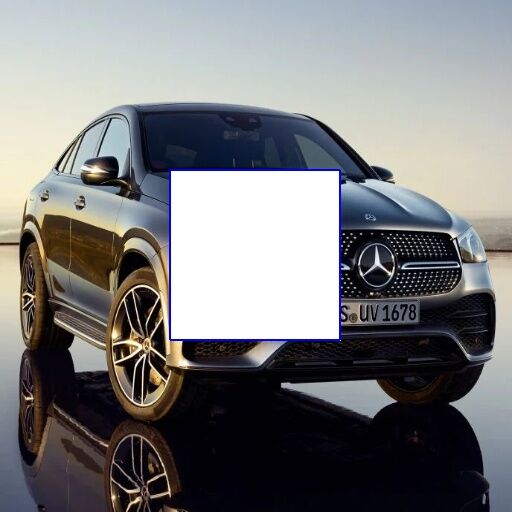}
    \end{subfigure}
    \begin{subfigure}{0.32 \columnwidth}
        \includegraphics[width=\textwidth]{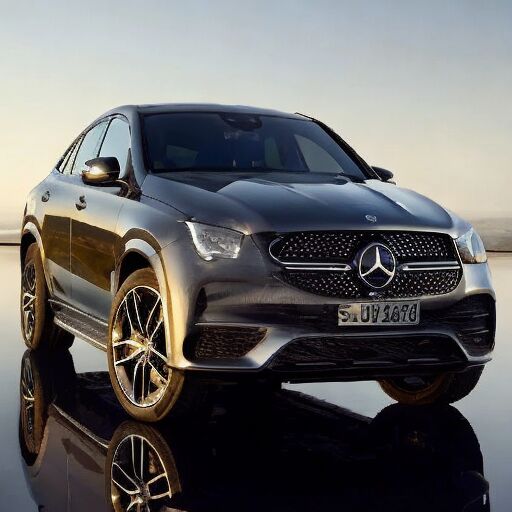}
    \end{subfigure}
    \begin{subfigure}{0.32 \columnwidth}
        \includegraphics[width=\textwidth]{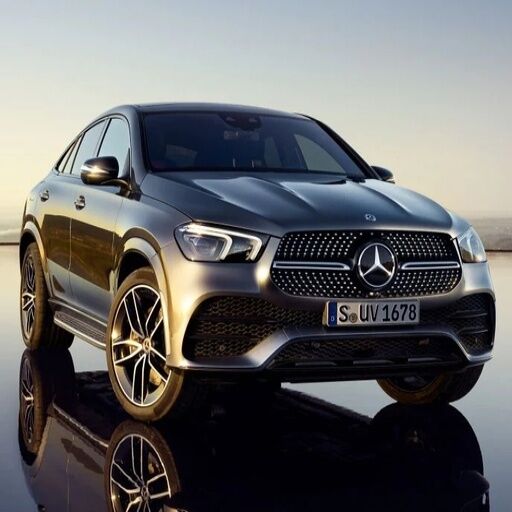}
    \end{subfigure}
    \\
    \begin{subfigure}{0.32 \columnwidth}
        \includegraphics[width=\textwidth]{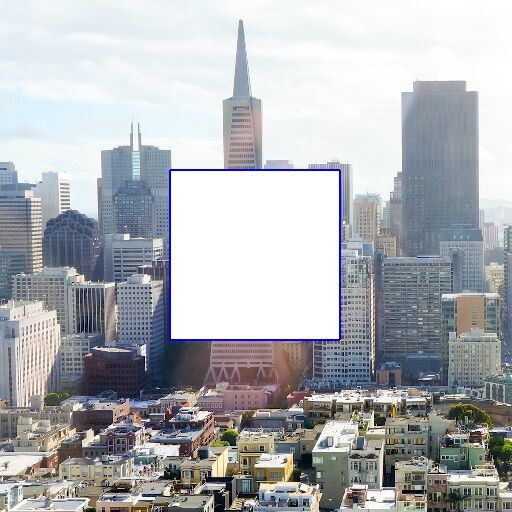}
        \caption{Input}
    \end{subfigure}
    \begin{subfigure}{0.32 \columnwidth}
        \includegraphics[width=\textwidth]{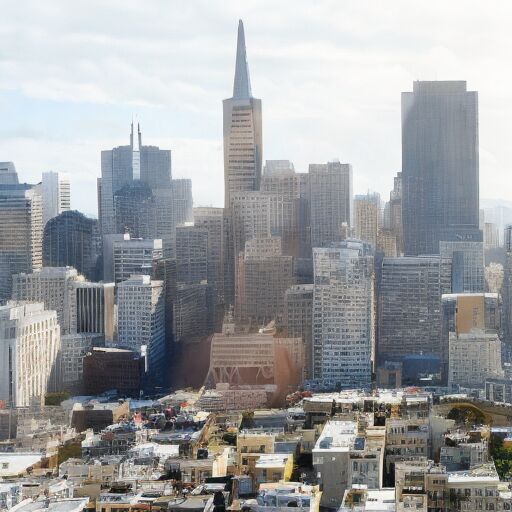}
        \caption{Output}
    \end{subfigure}
    \begin{subfigure}{0.32 \columnwidth}
        \includegraphics[width=\textwidth]{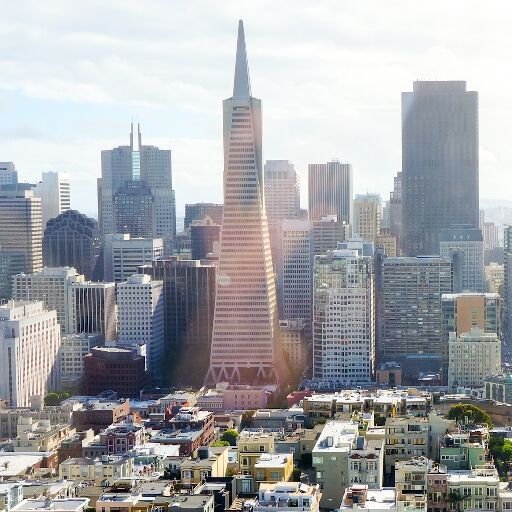}
        \caption{Ground truth}
    \end{subfigure}
    \caption{Visual results of our RefPaint Diffusion when applied to unconditional image inpainting. The blue bounding box indicates the inpainting mask.}
    \label{fig:blind}
\end{figure}

\begin{table}[t]
\centering
\resizebox{\columnwidth}{!}{%
\begin{tabular}{l|cc}
\toprule
\textbf{Model} & \textbf{CLIP w/ original} ($\uparrow$) & \textbf{CLIP w/ CP} ($\downarrow$) \\
\midrule
Copy-Paste (CP) & 0.2940 & 0.0\\
LaMa~\cite{suvorov2022resolution} & 0.1309 & 0.2676\\
SD Inpainting~\cite{rombach2022high} & 0.1392 & 0.2806\\
\textbf{Ours} & 0.2568 &  0.2586 \\ 
\bottomrule
\end{tabular}
}
\caption{Quantitative comparisons for reference-based painterly inpainting on random combinations of paintings from WikiArt~\cite{saleh2015large} and objects from COCO Captions~\cite{chen2015microsoft}.}
\label{tab:baselines}
\end{table}

\subsection{Qualitative Results}

\paragraph{Reference-based Painterly Inpainting}
\looseness=-1
Due to the space limit, we provide more visual comparisons in the supplementary material. As shown in Fig.~\ref{fig:results}, our proposed RefPaint Diffusion is able to generate diverse objects inside the masked regions even when both masks and reference objects are randomly sampled. Even when the reference object is complex (\textit{e.g.} the train) and has very different style information (\textit{e.g.} the flowers), our RefPaint Diffusion can apply the correct style information.  While maintaining semantic alignment with the reference objects, the overall image shows a consistent global style with the original artworks. 

\paragraph{General Reference-based Inpainting}
Our RefPaint Diffusion is out-of-the-box applicable to the general reference-based inpainting task. We experiment on the test data released by TransFill~\cite{zhou2021transfill}. As shown in Fig.~\ref{fig:transfill}, there is no domain gap between the reference image and the input image. Our RefPaint Diffusion is able to work well on this data and deliver visually pleasing results.

\paragraph{Unconditional Image Inpainting}
Thanks to the classifier-free guidance technique, our trained model can also be considered an unconditional model when setting $\omega=0$. And, not surprisingly, our RefPaint Diffusion can be applied to blind image inpainting. We experiment with the test data originally collected by Versatile Diffusion~\cite{vd} for image variation. As shown in Fig.~\ref{fig:blind}, when there is no reference image available, our model is able to perform unconditional image inpainting and still deliver photorealistic and coherent results.

\subsection{Limitations} Despite the encouraging visual results from RefPaint Diffusion, the model suffers from the slow inference speed of diffusion models, since multiple inferences are required during the reverse diffusion process. Moreover, challenging cases (e.g., multiple objects, objects with detailed textures) need to be further explored. In the future, we will investigate how to expand our method to more general scenarios.

\section{Conclusions}
\looseness=-1
In this work, we present a novel task named Reference-based Painterly Inpainting, where a novel real-world object is implanted into the artistic background image. We propose an effective RefPaint Diffusion framework based on the Versatile Diffusion~\cite{vd} backbone. We design a ladder-side branch and a masked fusion block for fine-tuning the model on our inpainting task. By manipulating the weights of classifier-free guidance at inference time, our model inherently supports tuning the strength of semantic alignment with reference objects and style alignment with background images.

\clearpage
\appendix

\begin{figure}
    \centering
    \includegraphics[width=0.75\columnwidth]{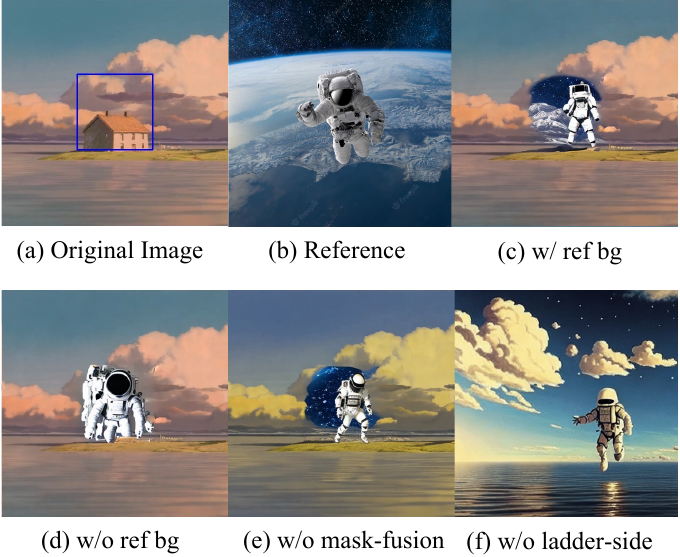}
    \caption{Clarification of Fig.~\ref{fig:teaser} (c-d)  and ablation  (e-f).}
    \label{fig:space}
\end{figure}
\begin{figure}
    \centering
    \includegraphics[width=0.8\columnwidth]{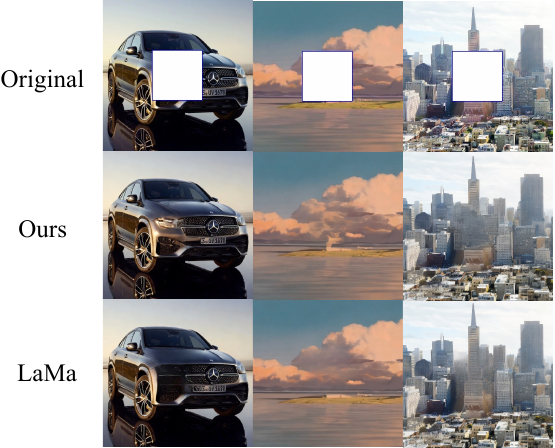}
    \caption{Compare with LaMa on blind inpainting.}
    \label{fig:lama}
\end{figure}
\begin{figure}
    \centering
    \includegraphics[width=0.8\columnwidth]{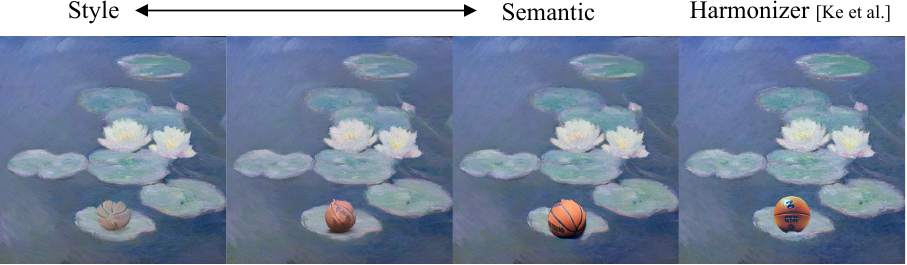}
    \caption{Compare with Joint Image Variation and Image Harmonization.}
    \label{fig:harmon}
\end{figure}

\begin{figure*}[p!]
    \vspace*{\fill}
    \centering
\begin{subfigure}{0.24 \textwidth}
\includegraphics[width=\textwidth]{images/origin/94fb08e3be3190ebff076996f7a2341ac_drawn.jpg}
\caption{Original Artwork}
\end{subfigure}
\begin{subfigure}{0.24\textwidth}
\includegraphics[width=\textwidth]{images/ref/8998_drawn.jpg}
\caption{Reference Object}
\end{subfigure}
\begin{subfigure}{0.24 \textwidth}
\includegraphics[width=\textwidth]{images/ours/cx_8998_x0_94fb08e3be3190ebff076996f7a2341ac_0.png_seed122_0.jpg}
\caption{Ours seed 0}
\end{subfigure}
\begin{subfigure}{0.24 \textwidth}
\includegraphics[width=\textwidth]{images/ours/cx_8998_x0_94fb08e3be3190ebff076996f7a2341ac_2.png_seed122_2.jpg}
\caption{Ours seed 1}
\end{subfigure}
\\
\begin{subfigure}{0.24 \textwidth}
\includegraphics[width=\textwidth]{images/ours/cx_8998_x0_94fb08e3be3190ebff076996f7a2341ac_3.png_seed122_3.jpg}
\caption{Ours seed 2}
\end{subfigure}
\begin{subfigure}{0.24 \textwidth}
\includegraphics[width=\textwidth]{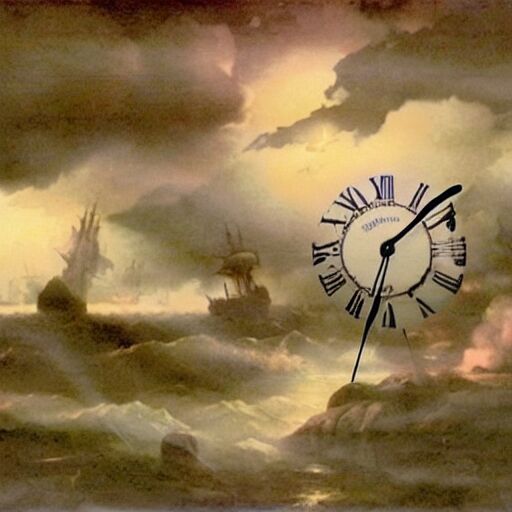}
\caption{SD seed 1}
\end{subfigure}
\begin{subfigure}{0.24 \textwidth}
\includegraphics[width=\textwidth]{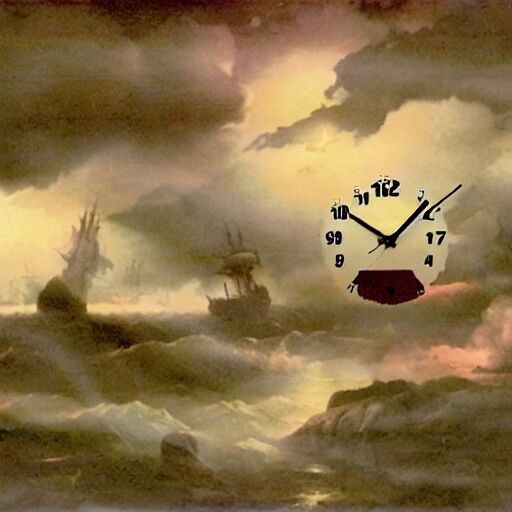}
\caption{SD seed 2}
\end{subfigure}
\begin{subfigure}{0.24 \textwidth}
\includegraphics[width=\textwidth]{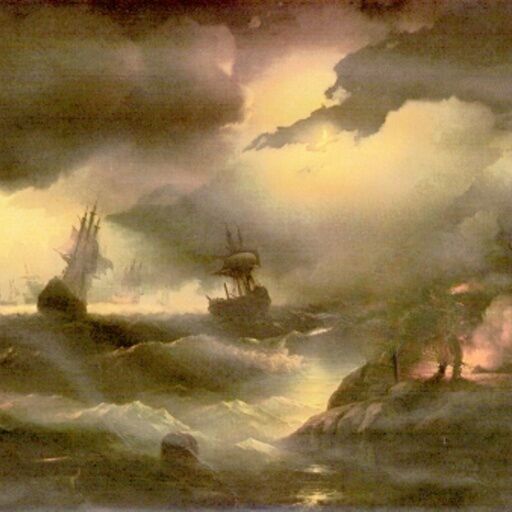}
\caption{LaMa}
\end{subfigure}
\\
\begin{subfigure}{0.24 \textwidth}
\includegraphics[width=\textwidth]{images/origin/353642766087c9c32f77460eacb34442c_drawn.jpg}
\caption{Original Artwork}
\end{subfigure}
\begin{subfigure}{0.24 \textwidth}
\includegraphics[width=\textwidth]{images/ref/74533_drawn.jpg}
\caption{Reference}
\end{subfigure}
\begin{subfigure}{0.24 \textwidth}
\includegraphics[width=\textwidth]{images/ours/cx_74533_x0_353642766087c9c32f77460eacb34442c_0.png_seed120_0.jpg}
\caption{Ours seed 0}
\end{subfigure}
\begin{subfigure}{0.24 \textwidth}
\includegraphics[width=\textwidth]{images/ours/cx_74533_x0_353642766087c9c32f77460eacb34442c_2.png_seed120_2.jpg}
\caption{Ours seed 1}
\end{subfigure}
\\
\begin{subfigure}{0.24 \textwidth}
\includegraphics[width=\textwidth]{images/ours/cx_74533_x0_353642766087c9c32f77460eacb34442c_3.png_seed120_3.jpg}
\caption{Ours seed 2}
\end{subfigure}
\begin{subfigure}{0.24 \textwidth}
\includegraphics[width=\textwidth]{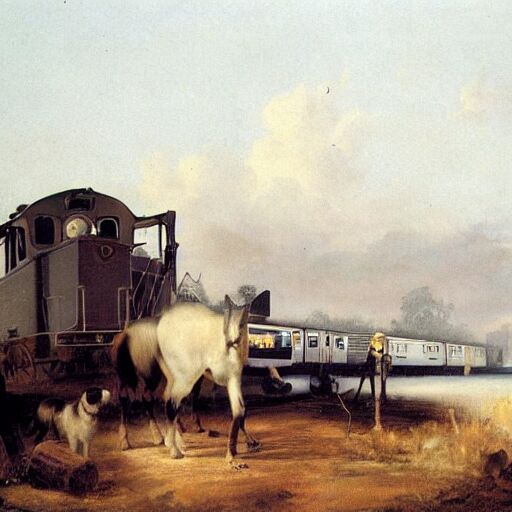}
\caption{SD seed 1}
\end{subfigure}
\begin{subfigure}{0.24 \textwidth}
\includegraphics[width=\textwidth]{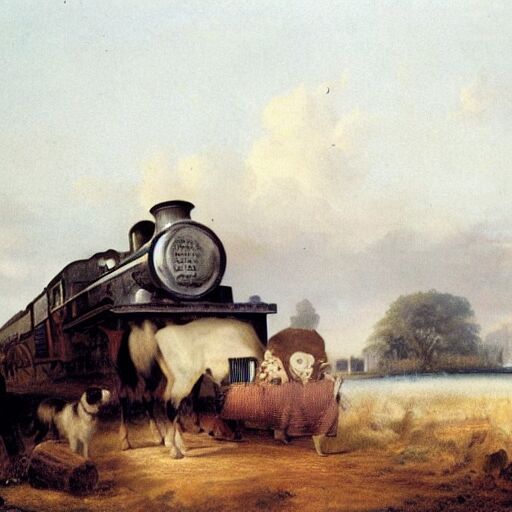}
\caption{SD seed 2}
\end{subfigure}
\begin{subfigure}{0.24 \textwidth}
\includegraphics[width=\textwidth]{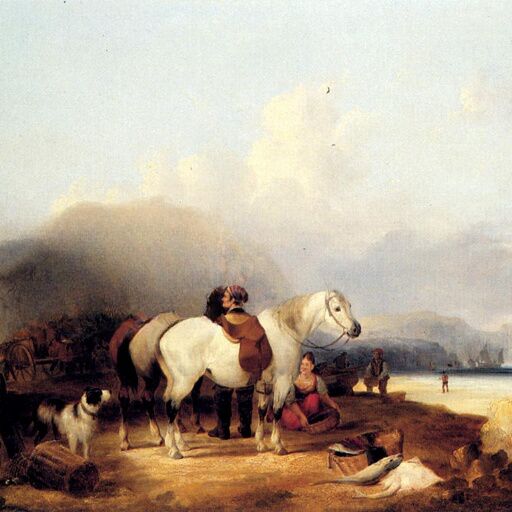}
\caption{LaMa}
\end{subfigure}
    \caption{Visual results for \underline{Ref}erence-based \underline{Paint}erly Inpainting results using \textbf{random} inpainting masks and \textbf{random} objects  from COCO Captions dataset. Blue bounding box represents the edited regions where we would like to inpaint. Red boundaries indicate the reference object. Text prompts for the objects are ``a photo of a small clock'' and ``a photo of a train on its tracks''.}
    \label{fig:results1}
\end{figure*}

\begin{figure*}
    \centering
\begin{subfigure}{0.24 \textwidth}
\includegraphics[width=\textwidth]{images/origin/ea88abb8440b62f68ed2996708fdd47fc_drawn.jpg}
\caption{Original Artwork}
\end{subfigure}
\begin{subfigure}{0.24 \textwidth}
\includegraphics[width=\textwidth]{images/ref/57242_drawn.jpg}
\caption{Reference Object}
\end{subfigure}
\begin{subfigure}{0.24 \textwidth}
\includegraphics[width=\textwidth]{images/ours/cx_57242_x0_ea88abb8440b62f68ed2996708fdd47fc_0.png_seed121_0.jpg}
\caption{Ours seed 0}
\end{subfigure}
\begin{subfigure}{0.24 \textwidth}
\includegraphics[width=\textwidth]{images/ours/cx_57242_x0_ea88abb8440b62f68ed2996708fdd47fc_2.png_seed121_2.jpg}
\caption{Ours seed 1}
\end{subfigure}
\\
\begin{subfigure}{0.24 \textwidth}
\includegraphics[width=\textwidth]{images/ours/cx_57242_x0_ea88abb8440b62f68ed2996708fdd47fc_3.png_seed121_3.jpg}
\caption{Ours seed 2}
\end{subfigure}
\begin{subfigure}{0.24 \textwidth}
\includegraphics[width=\textwidth]{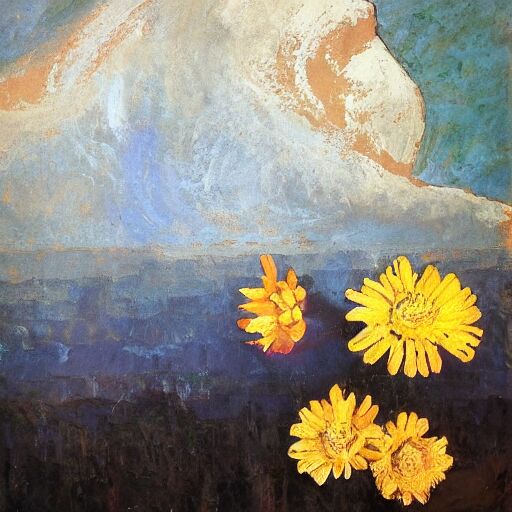}
\caption{SD seed 1}
\end{subfigure}
\begin{subfigure}{0.24 \textwidth}
\includegraphics[width=\textwidth]{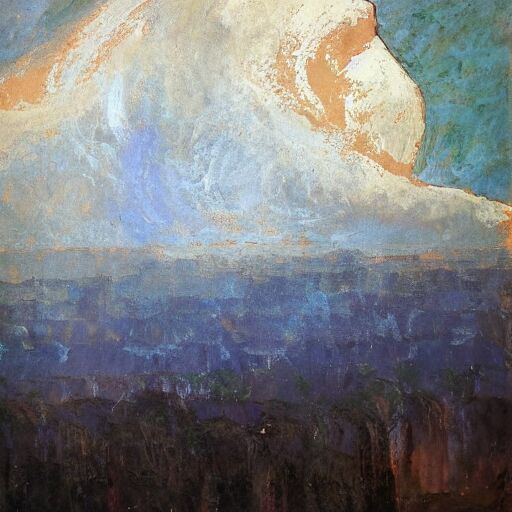}
\caption{SD seed 2}
\end{subfigure}
\begin{subfigure}{0.24 \textwidth}
\includegraphics[width=\textwidth]{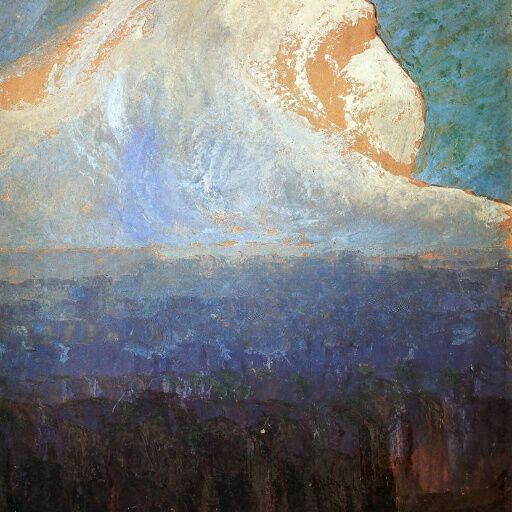}
\caption{LaMa}
\end{subfigure}
\\
\begin{subfigure}{0.24 \textwidth}
\includegraphics[width=\textwidth]{images/origin/6b54f9c0e754751b2e6014f0153b64bbc_drawn.jpg}
\caption{Original Artwork}
\end{subfigure}
\begin{subfigure}{0.24\textwidth}
\includegraphics[width=\textwidth]{images/ref/38310_drawn.jpg}
\caption{Reference Object}
\end{subfigure}
\begin{subfigure}{0.24 \textwidth}
\includegraphics[width=\textwidth]{images/ours/cx_38310_x0_6b54f9c0e754751b2e6014f0153b64bbc_0.png_seed124_0.jpg}
\caption{Ours seed 0}
\end{subfigure}
\begin{subfigure}{0.24 \textwidth}
\includegraphics[width=\textwidth]{images/ours/cx_38310_x0_6b54f9c0e754751b2e6014f0153b64bbc_2.png_seed124_2.jpg}
\caption{Ours seed 1}
\end{subfigure}
\\
\begin{subfigure}{0.24 \textwidth}
\includegraphics[width=\textwidth]{images/ours/cx_38310_x0_6b54f9c0e754751b2e6014f0153b64bbc_3.png_seed124_3.jpg}
\caption{Ours seed 2}
\end{subfigure}
\begin{subfigure}{0.24 \textwidth}
\includegraphics[width=\textwidth]{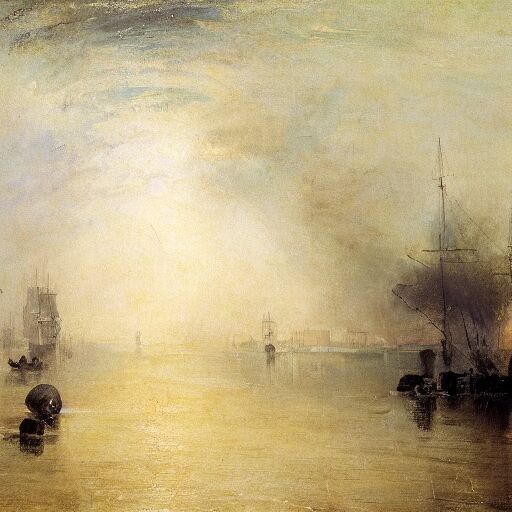}
\caption{SD seed 1}
\end{subfigure}
\begin{subfigure}{0.24 \textwidth}
\includegraphics[width=\textwidth]{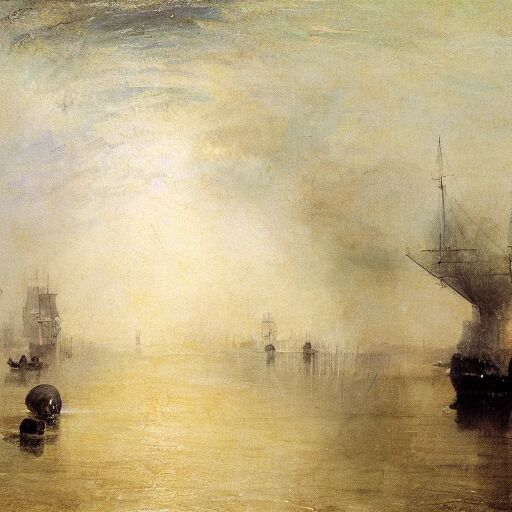}
\caption{SD seed 2}
\end{subfigure}
\begin{subfigure}{0.24 \textwidth}
\includegraphics[width=\textwidth]{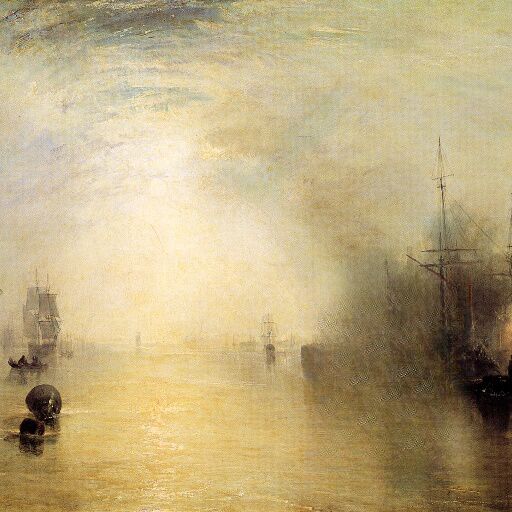}
\caption{LaMa}
\end{subfigure}
    \caption{Visual results for \underline{Ref}erence-based \underline{Paint}erly Inpainting results using \textbf{random} inpainting masks and \textbf{random} objects  from COCO Captions dataset. Blue bounding box represents the edited regions where we would like to inpaint. Red boundaries indicate the reference object. Text prompts for the objects are ``a photo of flowers'' and ``a photo of a large giraffe standing''.}
    \label{fig:results2}
\end{figure*}
\begin{figure*}
\begin{subfigure}{0.24 \textwidth}
\includegraphics[width=\textwidth]{images/origin/dfcb29a1633bbd3dbdb8083bcedd299ac_drawn.jpg}
\caption{Original Artwork}
\end{subfigure}
\begin{subfigure}{0.24\textwidth}
\includegraphics[width=\textwidth]{images/ref/29466_drawn.jpg}
\caption{Reference Object}
\end{subfigure}
\begin{subfigure}{0.24 \textwidth}
\includegraphics[width=\textwidth]{images/ours/cx_29466_x0_dfcb29a1633bbd3dbdb8083bcedd299ac_0.png_seed123_0.jpg}
\caption{Ours seed 0}
\end{subfigure}
\begin{subfigure}{0.24 \textwidth}
\includegraphics[width=\textwidth]{images/ours/cx_29466_x0_dfcb29a1633bbd3dbdb8083bcedd299ac_1.png_seed123_1.jpg}
\caption{Ours seed 1}
\end{subfigure}
\\
\begin{subfigure}{0.24 \textwidth}
\includegraphics[width=\textwidth]{images/ours/cx_29466_x0_dfcb29a1633bbd3dbdb8083bcedd299ac_3.png_seed123_3.jpg}
\caption{Ours seed 2}
\end{subfigure}
\begin{subfigure}{0.24 \textwidth}
\includegraphics[width=\textwidth]{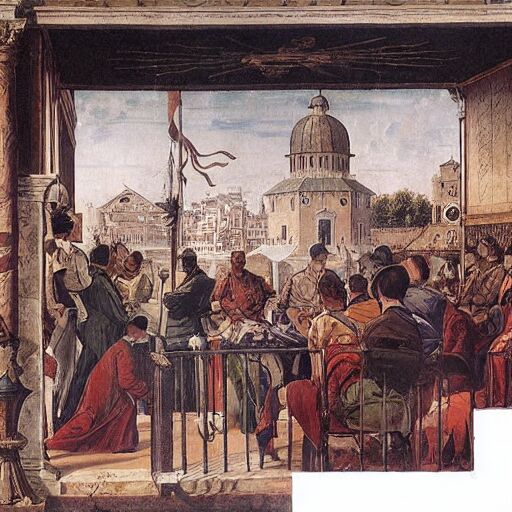}
\caption{SD seed 1}
\end{subfigure}
\begin{subfigure}{0.24 \textwidth}
\includegraphics[width=\textwidth]{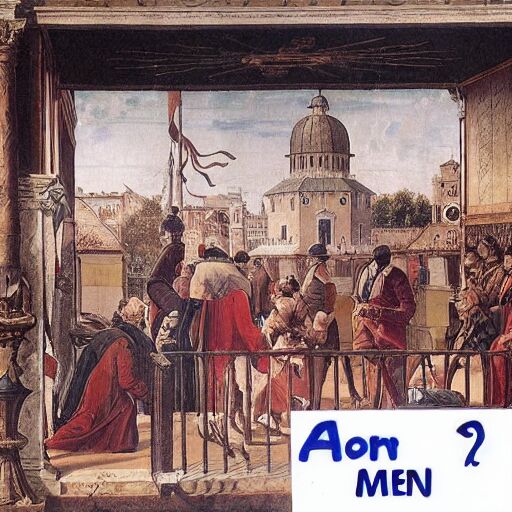}
\caption{SD seed 2}
\end{subfigure}
\begin{subfigure}{0.24 \textwidth}
\includegraphics[width=\textwidth]{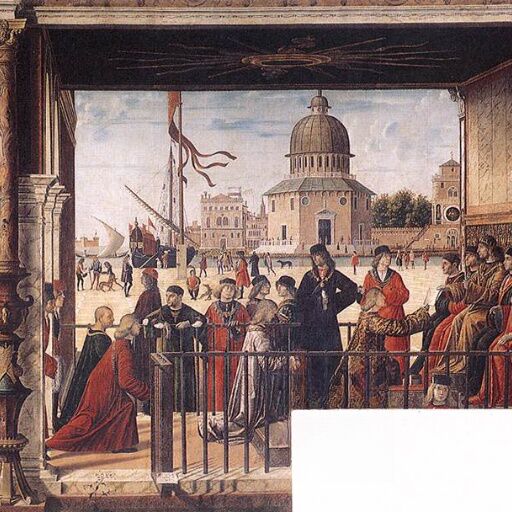}
\caption{LaMa}
\end{subfigure}
\\
\begin{subfigure}{0.24 \textwidth}
\includegraphics[width=\textwidth]{images/origin/7a2e8af80d678114c8538713a2969e88c_drawn.jpg}
\caption{Original Artwork}
\end{subfigure}
\begin{subfigure}{0.24\textwidth}
\includegraphics[width=\textwidth]{images/ref/439439_drawn.jpg}
\caption{Reference Object}
\end{subfigure}
\begin{subfigure}{0.24 \textwidth}
\includegraphics[width=\textwidth]{images/ours/cx_439439_x0_7a2e8af80d678114c8538713a2969e88c_0.png_seed124_0.jpg}
\caption{Ours seed 0}
\end{subfigure}
\begin{subfigure}{0.24 \textwidth}
\includegraphics[width=\textwidth]{images/ours/cx_439439_x0_7a2e8af80d678114c8538713a2969e88c_1.png_seed124_1.jpg}
\caption{Ours seed 1}
\end{subfigure}
\\
\begin{subfigure}{0.24 \textwidth}
\includegraphics[width=\textwidth]{images/ours/cx_439439_x0_7a2e8af80d678114c8538713a2969e88c_2.png_seed124_2.jpg}
\caption{Ours seed 2}
\end{subfigure}
\begin{subfigure}{0.24 \textwidth}
\includegraphics[width=\textwidth]{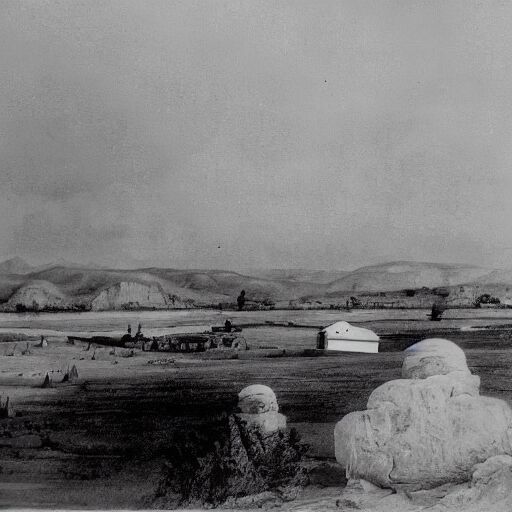}
\caption{SD seed 1}
\end{subfigure}
\begin{subfigure}{0.24 \textwidth}
\includegraphics[width=\textwidth]{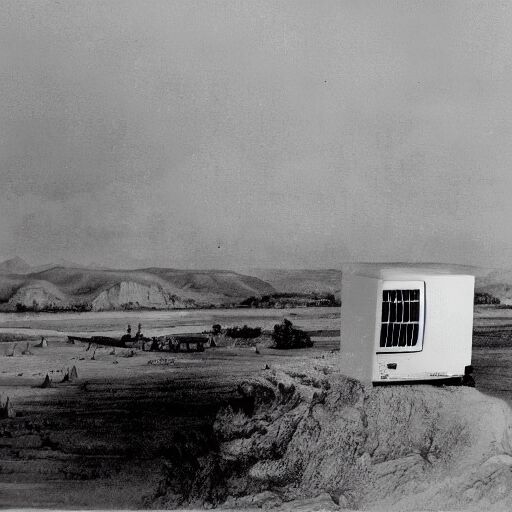}
\caption{SD seed 2}
\end{subfigure}
\begin{subfigure}{0.24 \textwidth}
\includegraphics[width=\textwidth]{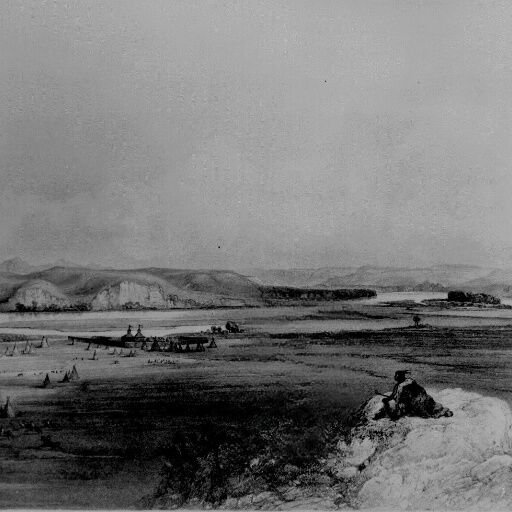}
\caption{LaMa}
\end{subfigure}
    \caption{Visual results for \underline{Ref}erence-based \underline{Paint}erly Inpainting results using \textbf{random} inpainting masks and \textbf{random} objects  from COCO Captions dataset. Blue bounding box represents the edited regions where we would like to inpaint. Red boundaries indicate the reference object. Text prompts for the objects are ``a photo of a boy'' and ``a photo of an open refrigerator full of items''.}
    \label{fig:results3}
    \vspace*{\fill}
\end{figure*}

\section{Comparison with Related Tasks}

We provide a thorough comparison between our proposed Reference-based Painterly Inpainting task and some related tasks in Tab.~\ref{tab:tasks}. Here we elaborate in detail.

\paragraph{Compare with Image Inpainting} Different from general blind image inpainting, our method supports input information from an additional reference image. Concretely speaking, blind image inpainting requires learning data-driven priors from a large-scale dataset. During testing, they don't have access to any additional information and the model is only allowed to utilize prior knowledge. One line of work, specifically studies reference-based image inpainting, where the model has access to a reference image with similar or almost identical content. However, unlike in our case where the reference image is real object and the background image is artistic, previous reference based image inpainting works focus on little to no domain gap. Specifically, existing methods learn pixel-level correspondence between the reference image and the input image using special architecture~\cite{zhou2021transfill} or geometric understanding~\cite{zhao2022geofill}. On the contrary, in our setting, there is hardly any correspondence available, considering the large domain gap.

\looseness=-1
\paragraph{Compare with Image Harmonization} Image harmonization usually handles the goal of direct image composition, which loyally pastes the reference object into the background image and harmonizes the overall visual quality. While there are works that study painterly harmonization that performs stylization to the objects to align with the overall style in images, the composited results usually look very close to the reference object if not identical. In comparison, our reference-based painterly inpainting is a more generative and creative task compared with immediate stitching. In other words, we seek variations to the reference object and want to generate similar objects instead of something identical. We never directly paste the reference object but try to capture the inherent semantics of the object and place semantically consistent contents in the region of interest.

\paragraph{Compare with Text-based Image Inpainting} The most related line of work to us is the text-based image inpainting implemented with recent text-to-image diffusion models. However, unlike in our case, where the reference information is an image, existing works handle reference information from text prompts. Text prompts describe the target object and contain less redundant information compared with a reference image. While more reference information is provided for the image setting, the generation is also limited to having an appearance similar to the reference image. In the text setting, on the other hand, the generated result is less constrained since it is only required to align with the text prompt.

\section{Additional Experimental Results}

\subsection{More Comparisons with Baseline Methods}
As shown in Fig.~\ref{fig:results1}, \ref{fig:results2}, \ref{fig:results3}, our results largely outperform the baseline methods. Specifically, SD inpainting tends to ignore the reference text prompt or paint outside the inpainting mask, leading to undesired overall results. LaMa, considering it's a blind inpainting method. inpaints a coherent background but doesn't work with additional reference information. These visual results also showcase that SD inpainting suffers from ignorance of the reference information and is not a perfect suit for our proposed reference-based painterly inpainting task. 

\subsection{Ablation Studies}
We have conducted two ablation studies in Tab.~\ref{tab:ablation} for the effectiveness of the ladder-side branch and the mask fusion blocks. Without the mask fusion blocks, the features from both branches are added directly, leading to color distortion in the background (Fig.~\ref{fig:space}(e)). When removing the ladder-side branch, the model performs global manipulations (Fig.~\ref{fig:space}(f)). 

\subsection{Comparison with Joint Image Variation and Image Harmonization.}
A naive combination of image variation (Versatile Diffusion~\cite{vd}) and image harmonization (Harmonizer~\cite{ke2022harmonizer}) models will suffer from the large domain gap between reference and background. As shown in Fig.~\ref{fig:harmon}, the inpainted object can be harmonized in appearance, but the resulting image has inconsistent style. On the contrary, our framework supports tuning the strength of reference semantics and background style, leading to coherent images.

\begin{table}[t]
\centering
\resizebox{0.8\columnwidth}{!}{%
\begin{tabular}{l|cc}
\toprule
\textbf{Model} & \textbf{CLIP w/ original} ($\uparrow$) & \textbf{CLIP w/ CP} ($\downarrow$) \\
\midrule
w/o ladder-side & 0.2219 & 0.2803\\
w/o mask-fusion & 0.2402 & 0.2701 \\
\textbf{Full Model} &\textbf{0.2568} &  \textbf{0.2586} \\ 
\bottomrule
\end{tabular}
}
\caption{Ablations. CP means copy-pasting the object.}
\label{tab:ablation}
\end{table}

{\small
\bibliographystyle{ieee_fullname}
\bibliography{egbib}

\begin{thebibliography}{10}\itemsep=-1pt

\bibitem{baek2016multiview}
Seung-Hwan Baek, Inchang Choi, and Min~H Kim.
\newblock Multiview image completion with space structure propagation.
\newblock In {\em Proceedings of the IEEE Conference on Computer Vision and Pattern Recognition}, pages 488--496, 2016.

\bibitem{barnes2009patchmatch}
Connelly Barnes, Eli Shechtman, Adam Finkelstein, and Dan~B Goldman.
\newblock Patchmatch: A randomized correspondence algorithm for structural image editing.
\newblock {\em ACM Trans. Graph.}, 28(3):24, 2009.

\bibitem{bertalmio2000image}
Marcelo Bertalmio, Guillermo Sapiro, Vincent Caselles, and Coloma Ballester.
\newblock Image inpainting.
\newblock In {\em Proceedings of the 27th annual conference on Computer graphics and interactive techniques}, pages 417--424, 2000.

\bibitem{bhavsar2010inpainting}
Arnav~V Bhavsar and Ambasamudram~N Rajagopalan.
\newblock Inpainting in multi-image stereo.
\newblock In {\em Joint Pattern Recognition Symposium}, pages 172--181. Springer, 2010.

\bibitem{chen2021crossvit}
Chun-Fu~Richard Chen, Quanfu Fan, and Rameswar Panda.
\newblock Crossvit: Cross-attention multi-scale vision transformer for image classification.
\newblock In {\em Proceedings of the IEEE/CVF international conference on computer vision}, pages 357--366, 2021.

\bibitem{chen2022diffusiondet}
Shoufa Chen, Peize Sun, Yibing Song, and Ping Luo.
\newblock Diffusiondet: Diffusion model for object detection.
\newblock {\em arXiv preprint arXiv:2211.09788}, 2022.

\bibitem{chen2015microsoft}
Xinlei Chen, Hao Fang, Tsung-Yi Lin, Ramakrishna Vedantam, Saurabh Gupta, Piotr Doll{\'a}r, and C~Lawrence Zitnick.
\newblock Microsoft coco captions: Data collection and evaluation server.
\newblock {\em arXiv preprint arXiv:1504.00325}, 2015.

\bibitem{cong2020dovenet}
Wenyan Cong, Jianfu Zhang, Li Niu, Liu Liu, Zhixin Ling, Weiyuan Li, and Liqing Zhang.
\newblock Dovenet: Deep image harmonization via domain verification.
\newblock In {\em Proceedings of the IEEE/CVF Conference on Computer Vision and Pattern Recognition}, pages 8394--8403, 2020.

\bibitem{cun2020improving}
Xiaodong Cun and Chi-Man Pun.
\newblock Improving the harmony of the composite image by spatial-separated attention module.
\newblock {\em IEEE Transactions on Image Processing}, 29:4759--4771, 2020.

\bibitem{dhariwal2021diffusion}
Prafulla Dhariwal and Alexander Nichol.
\newblock Diffusion models beat gans on image synthesis.
\newblock {\em Advances in Neural Information Processing Systems}, 34:8780--8794, 2021.

\bibitem{gu2023nerfdiff}
Jiatao Gu, Alex Trevithick, Kai-En Lin, Josh Susskind, Christian Theobalt, Lingjie Liu, and Ravi Ramamoorthi.
\newblock Nerfdiff: Single-image view synthesis with nerf-guided distillation from 3d-aware diffusion.
\newblock {\em arXiv preprint arXiv:2302.10109}, 2023.

\bibitem{ho2020denoising}
Jonathan Ho, Ajay Jain, and Pieter Abbeel.
\newblock Denoising diffusion probabilistic models.
\newblock {\em Advances in Neural Information Processing Systems}, 33:6840--6851, 2020.

\bibitem{ho2022classifier}
Jonathan Ho and Tim Salimans.
\newblock Classifier-free diffusion guidance.
\newblock {\em arXiv preprint arXiv:2207.12598}, 2022.

\bibitem{iizuka2017globally}
Satoshi Iizuka, Edgar Simo-Serra, and Hiroshi Ishikawa.
\newblock Globally and locally consistent image completion.
\newblock {\em ACM Transactions on Graphics (ToG)}, 36(4):1--14, 2017.

\bibitem{jia2006drag}
Jiaya Jia, Jian Sun, Chi-Keung Tang, and Heung-Yeung Shum.
\newblock Drag-and-drop pasting.
\newblock {\em ACM Transactions on graphics (TOG)}, 25(3):631--637, 2006.

\bibitem{jiang2021ssh}
Yifan Jiang, He Zhang, Jianming Zhang, Yilin Wang, Zhe Lin, Kalyan Sunkavalli, Simon Chen, Sohrab Amirghodsi, Sarah Kong, and Zhangyang Wang.
\newblock Ssh: a self-supervised framework for image harmonization.
\newblock In {\em Proceedings of the IEEE/CVF International Conference on Computer Vision}, pages 4832--4841, 2021.

\bibitem{ke2022harmonizer}
Zhanghan Ke, Chunyi Sun, Lei Zhu, Ke Xu, and Rynson~WH Lau.
\newblock Harmonizer: Learning to perform white-box image and video harmonization.
\newblock In {\em European Conference on Computer Vision}, pages 690--706. Springer, 2022.

\bibitem{kingma2021variational}
Diederik Kingma, Tim Salimans, Ben Poole, and Jonathan Ho.
\newblock Variational diffusion models.
\newblock {\em Advances in neural information processing systems}, 34:21696--21707, 2021.

\bibitem{li2022blip}
Junnan Li, Dongxu Li, Caiming Xiong, and Steven Hoi.
\newblock Blip: Bootstrapping language-image pre-training for unified vision-language understanding and generation.
\newblock In {\em International Conference on Machine Learning}, pages 12888--12900. PMLR, 2022.

\bibitem{liao2020guidance}
Liang Liao, Jing Xiao, Zheng Wang, Chia-Wen Lin, and Shin’ichi Satoh.
\newblock Guidance and evaluation: Semantic-aware image inpainting for mixed scenes.
\newblock In {\em Computer Vision--ECCV 2020: 16th European Conference, Glasgow, UK, August 23--28, 2020, Proceedings, Part XXVII 16}, pages 683--700. Springer, 2020.

\bibitem{liu2018image}
Guilin Liu, Fitsum~A Reda, Kevin~J Shih, Ting-Chun Wang, Andrew Tao, and Bryan Catanzaro.
\newblock Image inpainting for irregular holes using partial convolutions.
\newblock In {\em Proceedings of the European conference on computer vision (ECCV)}, pages 85--100, 2018.

\bibitem{luan2018deep}
Fujun Luan, Sylvain Paris, Eli Shechtman, and Kavita Bala.
\newblock Deep painterly harmonization.
\newblock In {\em Computer graphics forum}, volume~37, pages 95--106. Wiley Online Library, 2018.

\bibitem{lugmayr2022repaint}
Andreas Lugmayr, Martin Danelljan, Andres Romero, Fisher Yu, Radu Timofte, and Luc Van~Gool.
\newblock Repaint: Inpainting using denoising diffusion probabilistic models.
\newblock In {\em Proceedings of the IEEE/CVF Conference on Computer Vision and Pattern Recognition}, pages 11461--11471, 2022.

\bibitem{ma2021fov}
Liqian Ma, Stamatios Georgoulis, Xu Jia, and Luc Van~Gool.
\newblock Fov-net: Field-of-view extrapolation using self-attention and uncertainty.
\newblock {\em IEEE Robotics and Automation Letters}, 6(3):4321--4328, 2021.

\bibitem{ma2020learning}
Wei Ma, Mana Zheng, Wenguang Ma, Shibiao Xu, and Xiaopeng Zhang.
\newblock Learning across views for stereo image completion.
\newblock {\em IET Computer Vision}, 14(7):482--492, 2020.

\bibitem{nazeri2019edgeconnect}
Kamyar Nazeri, Eric Ng, Tony Joseph, Faisal~Z Qureshi, and Mehran Ebrahimi.
\newblock Edgeconnect: Generative image inpainting with adversarial edge learning.
\newblock {\em arXiv preprint arXiv:1901.00212}, 2019.

\bibitem{nichol2021glide}
Alex Nichol, Prafulla Dhariwal, Aditya Ramesh, Pranav Shyam, Pamela Mishkin, Bob McGrew, Ilya Sutskever, and Mark Chen.
\newblock Glide: Towards photorealistic image generation and editing with text-guided diffusion models.
\newblock {\em arXiv preprint arXiv:2112.10741}, 2021.

\bibitem{pathak2016context}
Deepak Pathak, Philipp Krahenbuhl, Jeff Donahue, Trevor Darrell, and Alexei~A Efros.
\newblock Context encoders: Feature learning by inpainting.
\newblock In {\em Proceedings of the IEEE conference on computer vision and pattern recognition}, pages 2536--2544, 2016.

\bibitem{peng2019element}
Hwai-Jin Peng, Chia-Ming Wang, and Yu-Chiang~Frank Wang.
\newblock Element-embedded style transfer networks for style harmonization.
\newblock In {\em BMVC}, page 201, 2019.

\bibitem{perez2003poisson}
Patrick P{\'e}rez, Michel Gangnet, and Andrew Blake.
\newblock Poisson image editing.
\newblock In {\em ACM SIGGRAPH 2003 Papers}, pages 313--318. 2003.

\bibitem{pitie2005n}
Francois Pitie, Anil~C Kokaram, and Rozenn Dahyot.
\newblock N-dimensional probability density function transfer and its application to color transfer.
\newblock In {\em Tenth IEEE International Conference on Computer Vision (ICCV'05) Volume 1}, volume~2, pages 1434--1439. IEEE, 2005.

\bibitem{poole2022dreamfusion}
Ben Poole, Ajay Jain, Jonathan~T Barron, and Ben Mildenhall.
\newblock Dreamfusion: Text-to-3d using 2d diffusion.
\newblock {\em arXiv preprint arXiv:2209.14988}, 2022.

\bibitem{ramesh2022hierarchical}
Aditya Ramesh, Prafulla Dhariwal, Alex Nichol, Casey Chu, and Mark Chen.
\newblock Hierarchical text-conditional image generation with clip latents.
\newblock {\em arXiv preprint arXiv:2204.06125}, 2022.

\bibitem{reinhard2001color}
Erik Reinhard, Michael Adhikhmin, Bruce Gooch, and Peter Shirley.
\newblock Color transfer between images.
\newblock {\em IEEE Computer graphics and applications}, 21(5):34--41, 2001.

\bibitem{ren2019structureflow}
Yurui Ren, Xiaoming Yu, Ruonan Zhang, Thomas~H Li, Shan Liu, and Ge Li.
\newblock Structureflow: Image inpainting via structure-aware appearance flow.
\newblock In {\em Proceedings of the IEEE/CVF International Conference on Computer Vision}, pages 181--190, 2019.

\bibitem{rombach2022high}
Robin Rombach, Andreas Blattmann, Dominik Lorenz, Patrick Esser, and Bj{\"o}rn Ommer.
\newblock High-resolution image synthesis with latent diffusion models.
\newblock In {\em Proceedings of the IEEE/CVF Conference on Computer Vision and Pattern Recognition}, pages 10684--10695, 2022.

\bibitem{saharia2022photorealistic}
Chitwan Saharia, William Chan, Saurabh Saxena, Lala Li, Jay Whang, Emily Denton, Seyed Kamyar~Seyed Ghasemipour, Burcu~Karagol Ayan, S~Sara Mahdavi, Rapha~Gontijo Lopes, et~al.
\newblock Photorealistic text-to-image diffusion models with deep language understanding.
\newblock {\em arXiv preprint arXiv:2205.11487}, 2022.

\bibitem{saleh2015large}
Babak Saleh and Ahmed Elgammal.
\newblock Large-scale classification of fine-art paintings: Learning the right metric on the right feature.
\newblock {\em arXiv preprint arXiv:1505.00855}, 2015.

\bibitem{salimans2022progressive}
Tim Salimans and Jonathan Ho.
\newblock Progressive distillation for fast sampling of diffusion models.
\newblock {\em arXiv preprint arXiv:2202.00512}, 2022.

\bibitem{saxena2023monocular}
Saurabh Saxena, Abhishek Kar, Mohammad Norouzi, and David~J Fleet.
\newblock Monocular depth estimation using diffusion models.
\newblock {\em arXiv preprint arXiv:2302.14816}, 2023.

\bibitem{schuhmann2022laion}
Christoph Schuhmann, Romain Beaumont, Richard Vencu, Cade Gordon, Ross Wightman, Mehdi Cherti, Theo Coombes, Aarush Katta, Clayton Mullis, Mitchell Wortsman, et~al.
\newblock Laion-5b: An open large-scale dataset for training next generation image-text models.
\newblock {\em arXiv preprint arXiv:2210.08402}, 2022.

\bibitem{singer2023text}
Uriel Singer, Shelly Sheynin, Adam Polyak, Oron Ashual, Iurii Makarov, Filippos Kokkinos, Naman Goyal, Andrea Vedaldi, Devi Parikh, Justin Johnson, et~al.
\newblock Text-to-4d dynamic scene generation.
\newblock {\em arXiv preprint arXiv:2301.11280}, 2023.

\bibitem{sohl2015deep}
Jascha Sohl-Dickstein, Eric Weiss, Niru Maheswaranathan, and Surya Ganguli.
\newblock Deep unsupervised learning using nonequilibrium thermodynamics.
\newblock In {\em International Conference on Machine Learning}, pages 2256--2265. PMLR, 2015.

\bibitem{song2020denoising}
Jiaming Song, Chenlin Meng, and Stefano Ermon.
\newblock Denoising diffusion implicit models.
\newblock {\em arXiv preprint arXiv:2010.02502}, 2020.

\bibitem{song2019generative}
Yang Song and Stefano Ermon.
\newblock Generative modeling by estimating gradients of the data distribution.
\newblock {\em Advances in neural information processing systems}, 32, 2019.

\bibitem{song2020score}
Yang Song, Jascha Sohl-Dickstein, Diederik~P Kingma, Abhishek Kumar, Stefano Ermon, and Ben Poole.
\newblock Score-based generative modeling through stochastic differential equations.
\newblock {\em arXiv preprint arXiv:2011.13456}, 2020.

\bibitem{song2018spg}
Yuhang Song, Chao Yang, Yeji Shen, Peng Wang, Qin Huang, and C-C~Jay Kuo.
\newblock Spg-net: Segmentation prediction and guidance network for image inpainting.
\newblock {\em arXiv preprint arXiv:1805.03356}, 2018.

\bibitem{suvorov2022resolution}
Roman Suvorov, Elizaveta Logacheva, Anton Mashikhin, Anastasia Remizova, Arsenii Ashukha, Aleksei Silvestrov, Naejin Kong, Harshith Goka, Kiwoong Park, and Victor Lempitsky.
\newblock Resolution-robust large mask inpainting with fourier convolutions.
\newblock In {\em Proceedings of the IEEE/CVF winter conference on applications of computer vision}, pages 2149--2159, 2022.

\bibitem{tao2010error}
Michael~W Tao, Micah~K Johnson, and Sylvain Paris.
\newblock Error-tolerant image compositing.
\newblock In {\em European Conference on Computer Vision}, pages 31--44. Springer, 2010.

\bibitem{thonat2016multi}
Theo Thonat, Eli Shechtman, Sylvain Paris, and George Drettakis.
\newblock Multi-view inpainting for image-based scene editing and rendering.
\newblock In {\em 2016 Fourth International Conference on 3D Vision (3DV)}, pages 351--359. IEEE, 2016.

\bibitem{vahdat2021score}
Arash Vahdat, Karsten Kreis, and Jan Kautz.
\newblock Score-based generative modeling in latent space.
\newblock {\em Advances in Neural Information Processing Systems}, 34:11287--11302, 2021.

\bibitem{wang2008stereoscopic}
Liang Wang, Hailin Jin, Ruigang Yang, and Minglun Gong.
\newblock Stereoscopic inpainting: Joint color and depth completion from stereo images.
\newblock In {\em 2008 IEEE Conference on Computer Vision and Pattern Recognition}, pages 1--8. IEEE, 2008.

\bibitem{watson2022novel}
Daniel Watson, William Chan, Ricardo Martin-Brualla, Jonathan Ho, Andrea Tagliasacchi, and Mohammad Norouzi.
\newblock Novel view synthesis with diffusion models.
\newblock {\em arXiv preprint arXiv:2210.04628}, 2022.

\bibitem{wexler2004space}
Yonatan Wexler, Eli Shechtman, and Michal Irani.
\newblock Space-time video completion.
\newblock In {\em Proceedings of the 2004 IEEE Computer Society Conference on Computer Vision and Pattern Recognition, 2004. CVPR 2004.}, volume~1, pages I--I. IEEE, 2004.

\bibitem{xu2022neurallift}
Dejia Xu, Yifan Jiang, Peihao Wang, Zhiwen Fan, Yi Wang, and Zhangyang Wang.
\newblock Neurallift-360: Lifting an in-the-wild 2d photo to a 3d object with 360° views.
\newblock {\em arXiv e-prints}, pages arXiv--2211, 2022.

\bibitem{vd}
Xingqian Xu, Zhangyang Wang, Eric Zhang, Kai Wang, and Humphrey Shi.
\newblock Versatile diffusion: Text, images and variations all in one diffusion model.
\newblock {\em arXiv preprint arXiv:2211.08332}, 2022.

\bibitem{yu2019free}
Jiahui Yu, Zhe Lin, Jimei Yang, Xiaohui Shen, Xin Lu, and Thomas~S Huang.
\newblock Free-form image inpainting with gated convolution.
\newblock In {\em Proceedings of the IEEE/CVF international conference on computer vision}, pages 4471--4480, 2019.

\bibitem{zhang2023adding}
Lvmin Zhang and Maneesh Agrawala.
\newblock Adding conditional control to text-to-image diffusion models.
\newblock {\em arXiv preprint arXiv:2302.05543}, 2023.

\bibitem{zhang2020deep}
Lingzhi Zhang, Tarmily Wen, and Jianbo Shi.
\newblock Deep image blending.
\newblock In {\em Proceedings of the IEEE/CVF Winter Conference on Applications of Computer Vision}, pages 231--240, 2020.

\bibitem{zhao2022geofill}
Yunhan Zhao, Connelly Barnes, Yuqian Zhou, Eli Shechtman, Sohrab Amirghodsi, and Charless Fowlkes.
\newblock Geofill: Reference-based image inpainting of scenes with complex geometry.
\newblock {\em arXiv preprint arXiv:2201.08131}, 2022.

\bibitem{zhou2021transfill}
Yuqian Zhou, Connelly Barnes, Eli Shechtman, and Sohrab Amirghodsi.
\newblock Transfill: Reference-guided image inpainting by merging multiple color and spatial transformations.
\newblock In {\em Proceedings of the IEEE/CVF Conference on Computer Vision and Pattern Recognition}, pages 2266--2276, 2021.

\bibitem{zhu2015learning}
Jun-Yan Zhu, Philipp Krahenbuhl, Eli Shechtman, and Alexei~A Efros.
\newblock Learning a discriminative model for the perception of realism in composite images.
\newblock In {\em Proceedings of the IEEE International Conference on Computer Vision}, pages 3943--3951, 2015.

\end{thebibliography}
}

\end{document}